\pdfoutput=1

\documentclass[11pt]{article}

\usepackage{acl}

\usepackage{times}
\usepackage{latexsym}

\usepackage[T1]{fontenc}

\usepackage[utf8]{inputenc}

\usepackage{microtype}
\usepackage{subcaption}
\usepackage{paralist, tabularx}

\usepackage{inconsolata}

\usepackage{arydshln}
\usepackage{graphicx}
\usepackage{multirow}
\usepackage{multicol}
\usepackage{hyperref}
\usepackage{url}
\usepackage{booktabs}
\usepackage{makecell}
\usepackage{xspace}
\newcommand{\ie}{\textit{i.e.}\xspace}

\newcommand{\frm}{\textbf{\textsc{InterIntent}}\xspace}

\title{\textsc{InterIntent}: Investigating Social Intelligence of LLMs via \\Intention Understanding in an Interactive Game Context}

\author{Ziyi Liu\thanks{~~Equal contribution.}\hspace{3mm} Abhishek Anand$^*$ \hspace{3mm} Pei Zhou\hspace{3mm} Jen-tse Huang\hspace{3mm}Jieyu Zhao \\
University of Southern California  \\
\small{\texttt{\{zliu2803, anandabh, peiz, jh\_116, jieyuz\}@usc.edu}}\\
}

\begin{document}
\maketitle

\begin{abstract}
Large language models (LLMs) have demonstrated the potential to mimic human social intelligence.
However, most studies focus on simplistic and static self-report or performance-based tests, which limits the depth and validity of the analysis.
In this paper, we developed a novel framework, \frm, to assess LLMs' social intelligence by mapping their ability to understand and manage intentions in a game setting. We focus on four dimensions of social intelligence: situational awareness, self-regulation, self-awareness, and theory of mind. Each dimension is linked to a specific game task: intention selection, intention following, intention summarization, and intention guessing. Our findings indicate that while LLMs exhibit high proficiency in selecting intentions, achieving an accuracy of 88\%, their ability to infer the intentions of others is significantly weaker, trailing human performance by 20\%. Additionally, game performance correlates with intention understanding, highlighting the importance of the four components towards success in this game. These findings underline the crucial role of intention understanding in evaluating LLMs' social intelligence and highlight the potential of using social deduction games as a complex testbed to enhance LLM evaluation. \frm contributes a structured approach to bridging the evaluation gap in social intelligence within multiplayer games.\footnote{Code is available at \href{https://github.com/uscnlp-lime/Inter-Intent}{https://github.com/uscnlp-lime/Inter-Intent}}

\end{abstract}

\section{Introduction}
\label{sec:intro}

The growing intelligence of large language models (LLMs) has facilitated diverse research on their capability to mimic human social intelligence~\citep{ziems2024large,dubova2022building,gweon2023socially,huang2023humanity,sap2022neural,wu2023coke}.
\textit{Social intelligence}, the ability to understand and manage one's own and others' actions and to act wisely in social relations~\cite{thorndike1920intelligence}, usually includes four key components~\cite{silvera2001tromso}:
(1) \textit{Situational Awareness}: The perception and comprehension of the elements in the environment~\cite{endsley1995toward}.
(2) \textit{Self-Regulation}: The process of guiding one's own thoughts, behaviors, and feelings to reach goals~\cite{bandura1991social}.
(3) \textit{Self-Awareness}: The understanding of one's own character, feelings, motives, and desires~\cite{gallup2003self}.
(4) \textit{Theory of Mind} (ToM): The knowledge about others' beliefs, intentions, and thoughts~\cite{baron1991precursors}.

\begin{figure*}[t]
\centering
\includegraphics[width=0.9\linewidth]{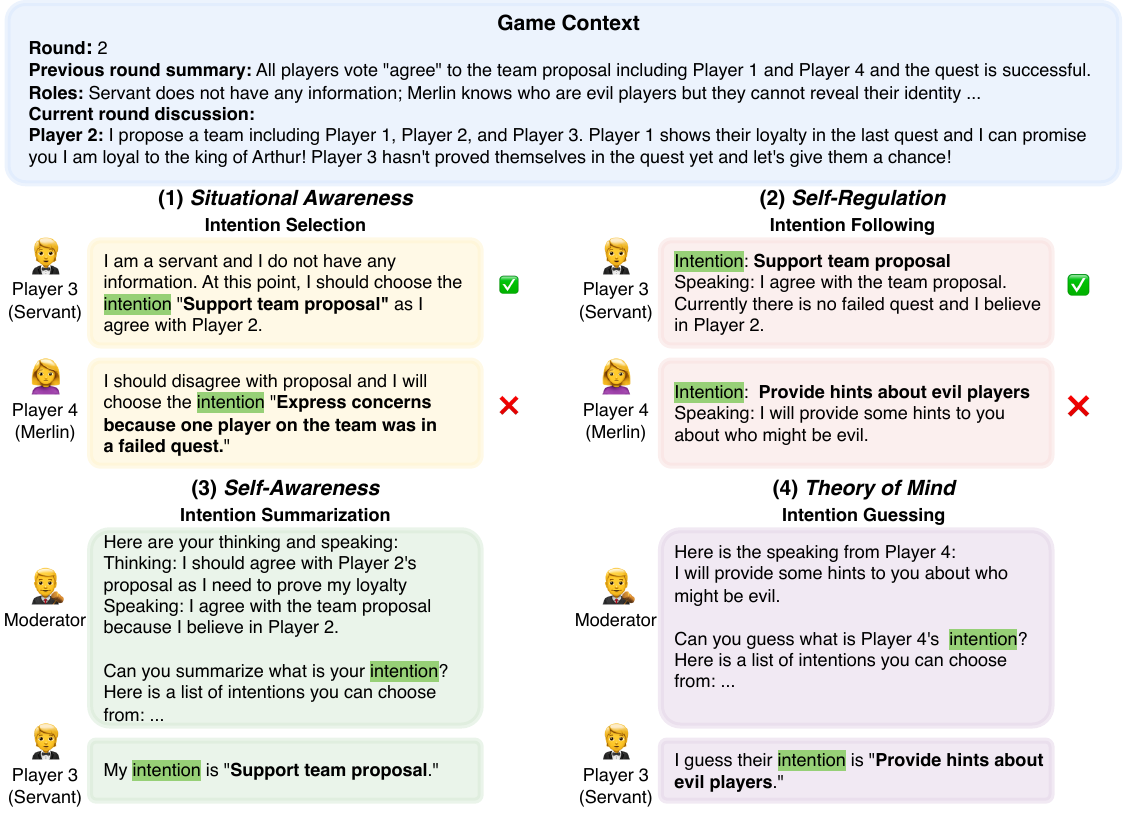}
\caption{
Four dimensions to assess social intelligence in Avalon. We provide a dynamic and complex gaming context for evaluations. For situational awareness, we provide both positive and negative examples. In the negative example, since the previous quest was successful and no player was in a failed quest, the intention is inappropriate as there was no failed quest. For self-regulation, we require models to provide explicit information rather than repeating the intentions. Intentions are in \textbf{bold} within the contexts.
}
\vspace{-0.3cm}
\label{fig:4_dimensions}
\end{figure*}

Despite adaptations for interactive social environments~\cite{zhou2023sotopia,zhou2023far,park2023generative}, most existing evaluations on LLMs~\cite{le2019revisiting,shapira2023clever} focus on straightforward and static daily scenarios without explicitly defined goals for them. To enhance the contexts with complexity and dynamics with clear goals for LLMs, we turn our attention to social deduction games.
These games require social interactions among players, providing a more diverse test bed for LLM evaluation~\cite{qiao2023gameeval}.
Existing studies have successfully utilized various games to analyze LLMs' social behaviors, such as deception and confrontation~\cite{liang2023leveraging,ibraheem2022putting,mansbach2021agent,meta2022human,o2023hoodwinked,xu2023exploring,xu2023language,wu2024enhance}.
However, these studies often focus on ad-hoc post-analysis of game performance and overlook systematic measurement of social intelligence, limiting the comprehensive understanding of LLMs' capabilities in social environments.

In this paper, we leverage one of the representatives in social deduction games, \textit{Avalon}~\cite{light2023avalonbench,wang2023avalons}, as the context for LLM evaluation.
Avalon is a social deduction game that relies on conversation, making it an ideal test bed with its goal-driven objectives and complex mechanisms, facilitating interactions among LLMs in a grounded environment.
We systematically evaluate LLMs' social intelligence across the four aforementioned dimensions.
Specifically, we propose to focus on players' \textit{Intentions}~\cite{malle2001intentions} at various phases of the game, since comprehension of one's and others' intentions plays a critical role in games~\cite{goodie2012levels}.
While previous research has explored intention detection~\citep{zhang2020intent,casanueva2020efficient}, the field remains largely underexplored in game contexts.
As shown in Figure~\ref{fig:4_dimensions}, we design four tests: (1) \textit{Intention Selection}, (2) \textit{Intention Following}, (3) \textit{Intention Summarization}, (4) \textit{Intention Guessing}, respectively for the four social intelligence dimensions.

\begin{figure*}[t]
\centering
\includegraphics[width=0.95\linewidth]{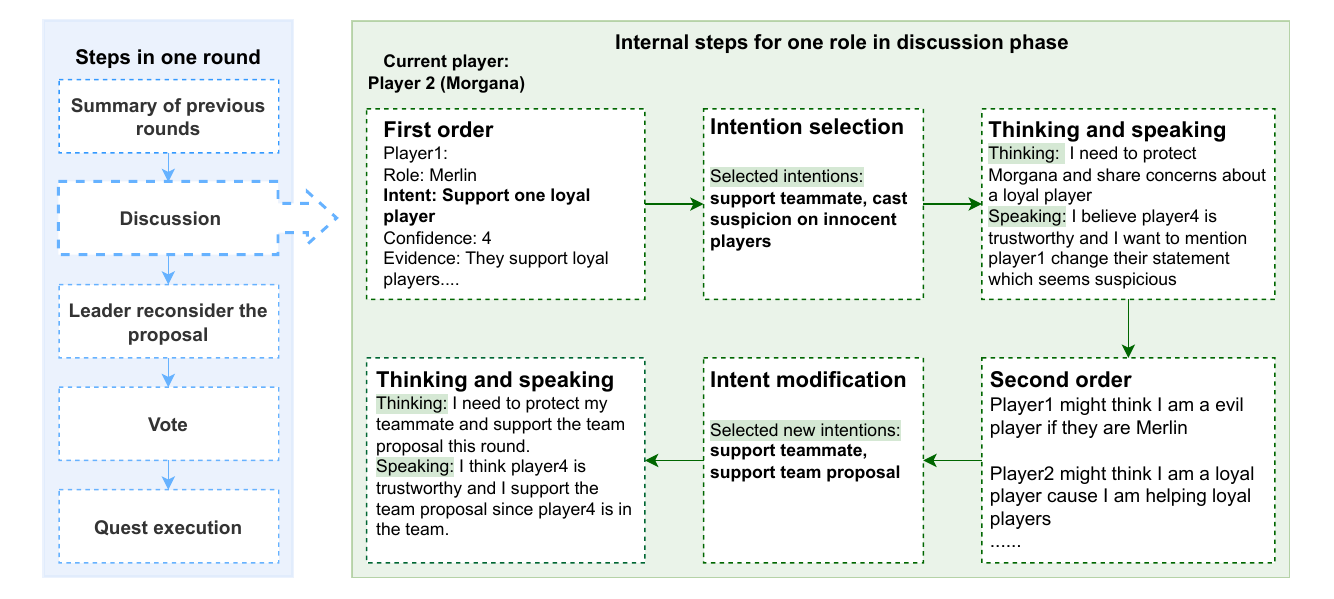}
\caption{The Avalon game process for one round. \textbf{Left}: the entire game pipeline. \textbf{Right}: the procedure for generating a single player's speech. }
\label{fig:game_procedure}
\vspace{-0.2cm}
\end{figure*}

We first develop a novel Avalon gaming framework enhanced with an intention-guided mechanism, drawing inspiration from previous studies~\cite{lan2023llm,shi2023cooperation,wang2023avalons,light2023text}.
Building on this, we introduced \frm, which is designed to dynamically generate contexts for evaluating social intelligence through LLMs' intention understanding.
With human annotation, our results reveal that GPT-3.5 and GPT-4 can select intentions correctly, with accuracies of 87.5\% and 88.8\%, respectively.
GPT-3.5 achieves a 69.5\% $F_1$ score in intention summarization, with GPT-4 reaching 83.8\%, surpassing human performance.
Furthermore, we find a strong correlation between high levels of intention comprehension and superior game win rate, highlighting the importance of the intention understanding components in enhancing LLMs' gaming proficiency.
However, the models fall short in intention following, with only 47.5\% and 64.8\% of responses rated as ``Good'' in human evaluations, and underperform in intention guessing, trailing human performance by 20\%.

Our contributions are summarized as follows:
\begin{compactitem}
    \item We bridge the gap between the evaluation of LLMs' social intelligence and LLM-based multiplayer social games.
    \item We introduce \frm, a framework designed to systematically analyze social intelligence through players' intention understanding within the Avalon game environment.
    \item Extensive experiments on various LLMs reveal their underperformance in intention guessing against human performance.
\end{compactitem}
\section{Intention-Guided Avalon Framework}

We begin by outlining the entire game pipeline before focusing on the mechanics of intention-guided gameplay, which includes the categorization and application of intentions throughout the game.

\subsection{LLM-Based Avalon Game}

Avalon is a social deduction game where players are either \textit{loyal} or \textit{evil}, aiming to succeed in or sabotage quests through strategy, persuasion, deduction, and deception.
The primary goal is for loyal players to complete quests, while evil players aim to fail them.\footnote{The detailed rules of Avalon can be found in \url{https://avalon.fun/pdfs/rules.pdf}.}
Leveraging the publicly available prompts proposed by~\citet{wang2023avalons}, which includes first-order and second-order reasoning before speaking, we build a framework allowing five to ten players in gameplay (we set the players number as five in this paper which includes 2 evil players and 3 loyal players).
As shown in Figure~\ref{fig:game_procedure}, each game round consists of five principal components: summarizing previous rounds, discussing strategies, reconsidering the team proposal by the leader, voting, and executing the quest.
Notably, we introduce an additional component, the leader's reconsideration phase as a critical, previously unexplored component in Avalon literature.

\subsection{Intention Categorization}

\begin{table*}
\resizebox{1.0\linewidth}{!}{
    \begin{tabular}{cll}
    \toprule
    \textbf{Score} & \multicolumn{1}{c}{\bf Criterion} & \multicolumn{1}{c}{\bf Example} \\
    \midrule
    1 & \makecell[l]{The content does not mention the intentions at all. }&\makecell[l]{I am a loyal player and I support this proposal.}\\
    \hdashline
    2 & \makecell[l]{The content simply copies and pastes intentions.}&\makecell[l]{I express concerns about a player from a failed quest.}\\
    \hdashline
    3 & \makecell[l]{The content follows the intentions but has wrong \\context information.}&\makecell[l]{I suggest the leader should not include player3 in\\ the team as they failed the quest before. \\(\textcolor{blue}{The context is the team does not include Player 3})}\\
    \hdashline
    4 & \makecell[l]{The content follows the intentions but lacks \\useful information.}&\makecell[l]{I think some players should not be in the team as they \\in a failed team before. \\ (\textcolor{blue}{Lack of information of which player})}\\
    \hdashline
    5 & \makecell[l]{The content follows the intentions well with\\ clear information if required. }&\makecell[l]{I suggest the leader reconsider including Player 1 \\as they are in a failed quest so we cannot be sure \\of their loyalty.}\\
    \bottomrule
    \end{tabular}
}
\caption{Criteria for annotating intention following (speaking), accompanied by examples. The \textbf{intention} used across examples is ``Express concerns about a player from a failed quest team and suggest to not include them in the current team.'' The \textbf{context} is ``The team proposal is Player 1 and Player 2, with Player 1 being on a failed quest.'' We consider the score of ``including wrong context knowledge'' higher than ``copy and paste'' because we focus on whether the intention following has an impact on the game.}
\label{tab:score_example}
\end{table*}

To study intentions within Avalon gameplay systematically, we adopt the definition by \citet{kennington2022understanding}, viewing intention as a choice coupled with commitment.
This perspective expands intention beyond a mere mental state to include a discernible commitment to act with purpose.
In line with \citet{wang2023avalons} and \citet{xu2023exploring}, we identify seven categories of intentions in LLMs: \textit{Interrogation, Defense, Confrontation, Concealment, Deception, Persuasion}, and \textit{Teamwork}.
Initially, we derive a list of intentions from our direct experiences with the game of Avalon.
We then utilize GPT-3.5 to extend these initial categorizations.
Human annotators subsequently refine this expanded list, removing any intentions deemed unreasonable or redundant, thus producing a concise set of relevant intentions.
This refinement is iterative, resulting in the final enumeration detailed in Appendix Sec.~\ref{sec:app:intentions}.

\subsection{Intention-Guided Game Playing}

We introduce an intention-guided gameplay mechanism that integrates our defined intentions within the game environment, offering two main benefits.
First, it enhances the performance of LLMs by focusing on explicit intention discussions.
Second, it facilitates the evaluation of the four social intelligence components.

As depicted in Figure~\ref{fig:game_procedure}, the discussion phase involves several key steps to facilitate intentional interaction among players.
Initially, players use a first-order prompt for deductive reasoning about each player's role and intentions.
Subsequently, players select two or three intentions from a predefined set according to the current game context.
This selection process guides players in generating their thoughts and statements based on the chosen intentions.
Players then engage in second-order reasoning to evaluate how their statements might be interpreted by others, allowing them to reconsider and adjust their initial intentions.
Finally, players express their refined thoughts and statements during the game's discussion segment, enhanced by this iterative reflection process. Full prompts are shown in Appendix Sec.~\ref{sec:app:game_prompts}.
\section{Intention-Centric Evaluations}

This section introduces our evaluations, designed around the concept of intentions, and discusses their correlation with the four components of social intelligence outlined in Sec.~\ref{sec:intro}.
Intention is fundamental to social intelligence, essential for effective communication~\cite{tomasello2023having}, strategic influence, adaptation to dynamic social interactions~\cite{zelazo2023developing}, and achieving desired outcomes by helping us interpret and predict behaviors~\cite{yott2016infants}.
Our evaluation order corresponds to the game-playing process.

\subsection{Situational Awareness: Intention Selection}

Situational awareness is the ability to perceive environmental elements over time and space, understand their significance, and predict their future status~\cite{li2024social,endsley1995toward}.
In our study, we assess the situational awareness of LLMs in social contexts by evaluating their intention selection.
LLMs are requested to select intentions based on summarization, first-order reasoning, and the ongoing dialogue in the current round.
Ideally, LLMs should exhibit a sharp awareness of dynamic interactions, choosing intentions that are contextually appropriate rather than contradictory.
We deem intentions unreasonable if they are inconsistent with:
\begin{compactitem}
    \item Established facts—for instance, when all players have voted in the previous proposal, a player intends to question why someone did not vote.
    \item Role profiles—such as an evil player forgetting their identity and intending to mistakenly support the loyal side.
    \item Other intentions—like a player intends to play Merlin and Percival simultaneously.
\end{compactitem}
We use binary indicators to evaluate the reasonableness of intentions, assigning 1 to reasonable intentions and 0 to unreasonable ones. Previous hallucinations, such as manipulated information, might influence the current player's decisions. However, we maintain strict criteria, as we expect the player to be capable of recognizing these hallucinations.

\subsection{Self-Regulation: Intention Following}

Self-regulation involves guiding one’s own thoughts, behaviors, and feelings to achieve goals, thereby requiring individuals to contribute to their own motivation~\cite{fitzsimons2010interpersonal}.
In our study, we evaluate the self-regulation abilities of LLMs by assessing their adherence to selected intentions, which includes two main perspectives: thinking (planning) and speaking (implementing), as depicted in Figure~\ref{fig:game_procedure}.
Due to the abstract nature of the thinking phase, our criteria are lenient: we consider the LLM's thought process as correct if it reflects the intended goal.
While we do not demand informativeness in the thinking phase, we require validity and penalize scores for hallucinations or omission of intentions.
In contrast, for the speaking phase, we require the models to execute actions informatively and without hallucinations for a response to be considered good.
We employ a Likert scale to annotate outcomes for both phases, with scores ranging from 5 (completely correct) to 1 (completely incorrect), with 3 representing a borderline.
Table~\ref{tab:score_example} provides detailed criteria and examples for the annotation.

\subsection{Self-Awareness: Intention Summarization}

Self-awareness refers to an individual's understanding of their character, emotions, motives, and desires~\cite{gallup2003self}.
Our evaluation investigates the capability of LLMs to accurately identify their own intentions through analysis of their internal thought processes and speeches in the current round.
This evaluation serves as the converse to intention following.
LLMs are expected not only to execute intentions precisely but also to articulate their underlying motivations.

\subsection{Theory of Mind: Intention Guessing}

ToM involves understanding others by attributing mental states to them~\cite{kosinski2023theory}.
Previous evaluations of LLMs' ToM focused on scenarios where models are provided with complete contexts for interpreting characters’ mental states~\cite{gandhi2024understanding}.
Our evaluation raises the complexity and challenge for LLMs by providing only limited information, simulating real-world conditions.
Specifically, LLMs are required to deduce players' intentions from their speeches alone, representing a more rigorous test of their capability to comprehend and anticipate mental states.
\section{Experimental Settings and Results}

We run 40 games using gpt-3.5-turbo-1106 and 5 games using gpt-4-1106, with details in Appendix Sec.~\ref{sec:app:models}.
Sample size statistics across all experiments are presented in Table~\ref{tab:statistics}, and game performance metrics are discussed in Appendix Sec.~\ref{sec:app:game}.

\begin{table}
\centering
\scalebox{0.8}{
\begin{tabular}{lrr}
\toprule
\textbf{Models}&GPT-3.5&GPT-4\\
\midrule
\textbf{Intention Selection}&2,440&350\\
\textbf{Intention Following}&7,507&1,278\\
\textbf{Intention Summarization}&2,316&261\\
\textbf{Intention Guessing}&2,235&201\\
\bottomrule
\end{tabular}
}
\caption{Statistics for sample size over all experiments. Intention following includes thinking and speaking. }
\vspace{-0.3cm}
\label{tab:statistics}
\end{table}

\subsection{Human Annotation}

\paragraph{Intention Selection and Following}

\begin{figure*}
    \centering
    \includegraphics[width=\textwidth]{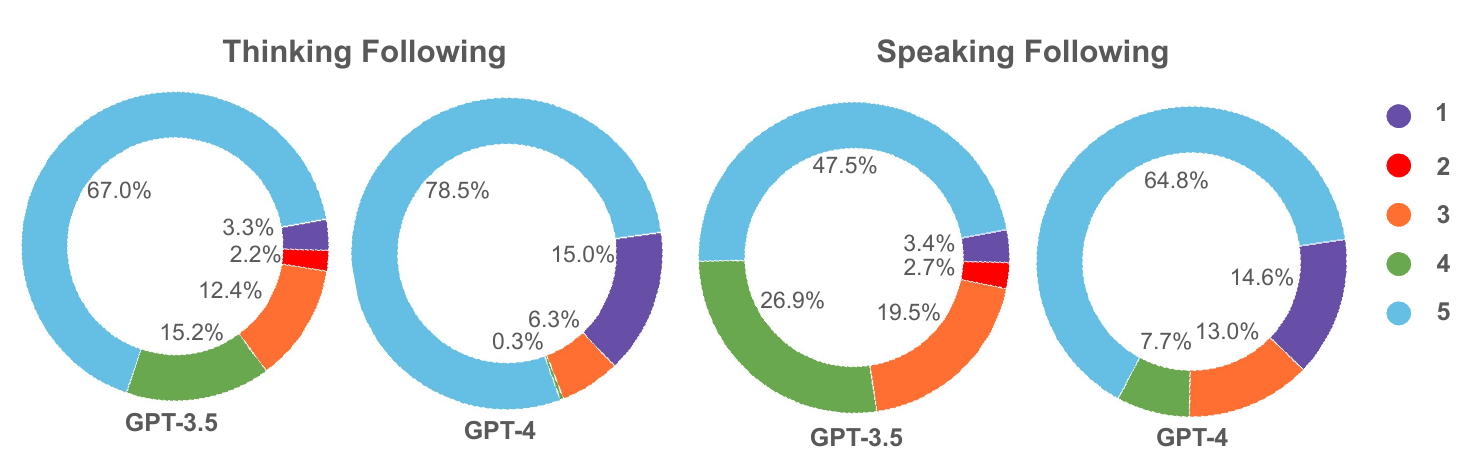}
    \caption{Self-regulation results. The results show the percentage of each score over all data samples. Scores 1-5 are evaluation criteria (Table \ref{tab:score_example}). Score 5 means the best while score 1 means the worst. 
    }
    \vspace{-0.2cm}
    \label{fig:following}
    
\end{figure*}

Figure~\ref{fig:4_dimensions} illustrates that intention selection and following necessitate extensive human annotation.
We obtain a total of 6,316 intentions in the 40 games utilizing GPT-3.5, presenting a significant workload for annotators who are required to meticulously examine every thinking and each speaking result.
Therefore, we first annotate intentions from five games to identify those with significant, either positively or negatively, impacts on game outcomes.
These ``impactful intentions'' are then prioritized for further study.
The methodology for selecting these intentions and the correlation between intention understanding and game performance is detailed in Appendix Sec.~\ref{app:intention_game}.
This refinement process isolates 2,440 intentions from GPT-3.5 and 350 intentions from GPT-4 for annotation (Table \ref{tab:statistics}).  

We recruit five computer science master's students for annotation, each assigned seven to eight game contexts.\footnote{Hourly payment for all annotators in this work is \$16.} 
Unimpactful intentions within each context are masked to maintain focus.
To calculate inter-rater reliability, two files are commonly assigned across all annotators.
The Fleiss' Kappa~\cite{fleiss1971measuring} is calculated pairwise, with averages derived across all pairs.
For the intention following, we group the scores into two categories: 1$\sim$3 and 4$\sim$5. Scores of 5 represent a perfect response, while 4 indicates a nearly perfect response. Scores from 1 to 3 signify that the response contains significant flaws.
The scores are shown in Table~\ref{tab:agreement_score}. Additional grouping results are provided in the Appendix Sec  \ref{sec:app:anno}.

\begin{table}
\centering
\scalebox{0.80}{
\begin{tabular}{cccc}
\toprule
& \makecell{Intention \\ Selection} & \makecell{Intention \\ Following \\ (Thinking)} & \makecell{Intention \\ Following \\ (Speaking)}\\
\midrule
Fleiss' Kappa & $0.69_{\pm 0.08}$ & $0.54_{\pm 0.06}$ & $0.49_{\pm0.06}$\\
\bottomrule
\end{tabular}
}
\caption{Inter-rater agreement score among human annotators. We show mean $\pm$ standard deviation over pairs of annotators.}
\vspace{-0.3cm}
\label{tab:agreement_score}
\end{table}

\paragraph{Intention Summarization and Guessing}

Evaluations of intention summarization and guessing are separated from the gameplay pipeline because as interactions progress, the clarity of LLMs' outputs often deteriorates rather than improves.
At the game's end, we extract the contexts including roles, discussions, voting outcomes, and quest results, in a structured format.
Notably, intention summarization and guessing are not critical to the game’s core mechanics.
By separating these elements, we can still ensure the comprehensiveness of structured game information, as detailed in Appendix Sec.~\ref{sec:appendix:structured_con}.

Summarizing and guessing intentions do not require human annotation, as we can directly compare the outcomes with players' earlier choices.
However, evaluating the performance of LLMs against humans remains essential.
Unlike the straightforward tasks of intention selection and following, which generally require the exclusion of incorrect information—a task in which humans excel—summarizing and guessing intentions demands more social intelligence, a more complex human capability.
We conduct a user study with three master's students who answer a total of 300 questions, 200 for GPT-3.5, and 100 for GPT-4.


\begin{figure*}
  \centering
  \begin{subfigure}[b]{0.32\textwidth}
    \centering
    \includegraphics[width=\textwidth]{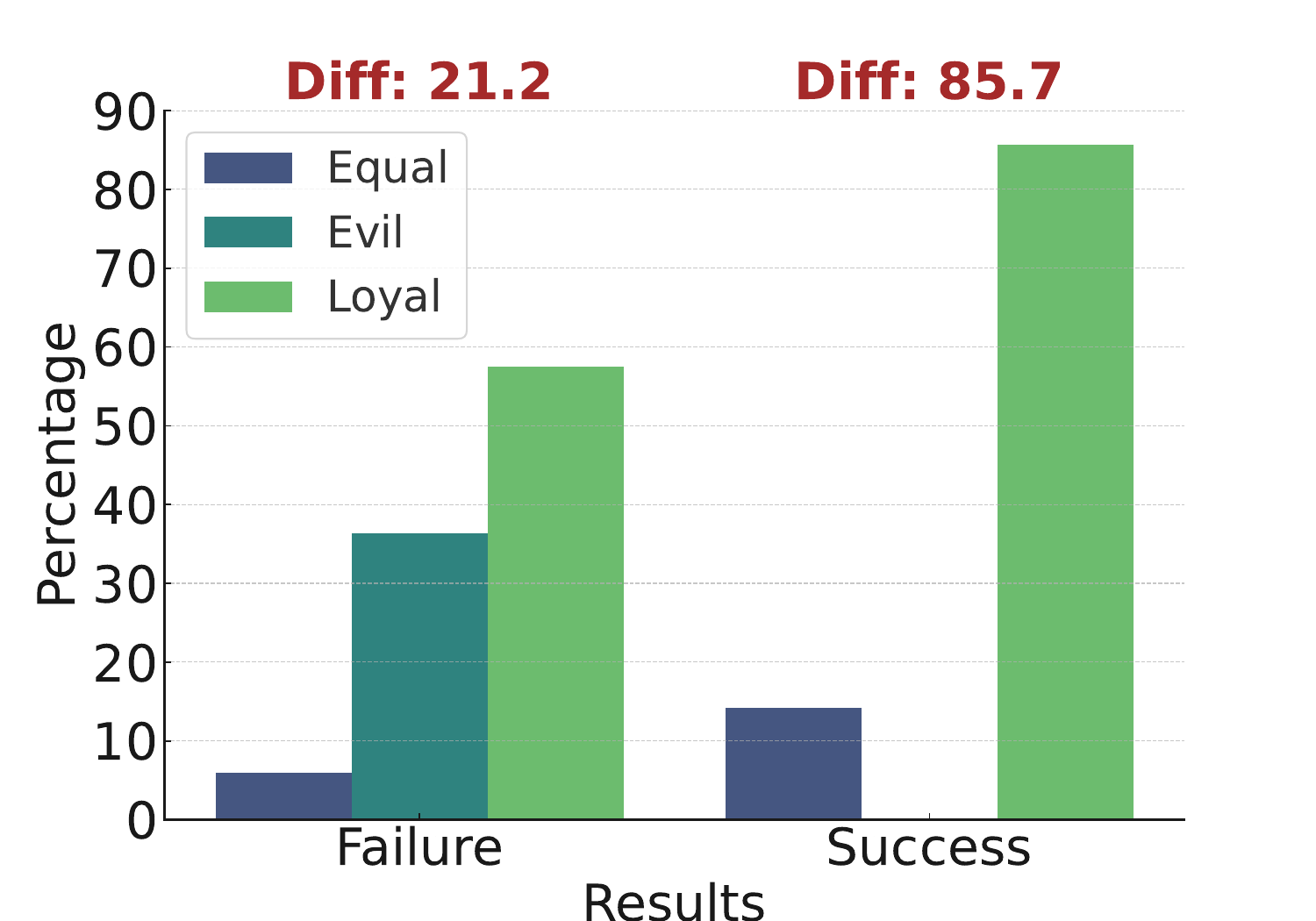}
    \caption{Intention Selection (Game Wise)}
    \label{fig:image1}
  \end{subfigure}
  \hfill
  \begin{subfigure}[b]{0.32\textwidth}
    \centering
    \includegraphics[width=\textwidth]{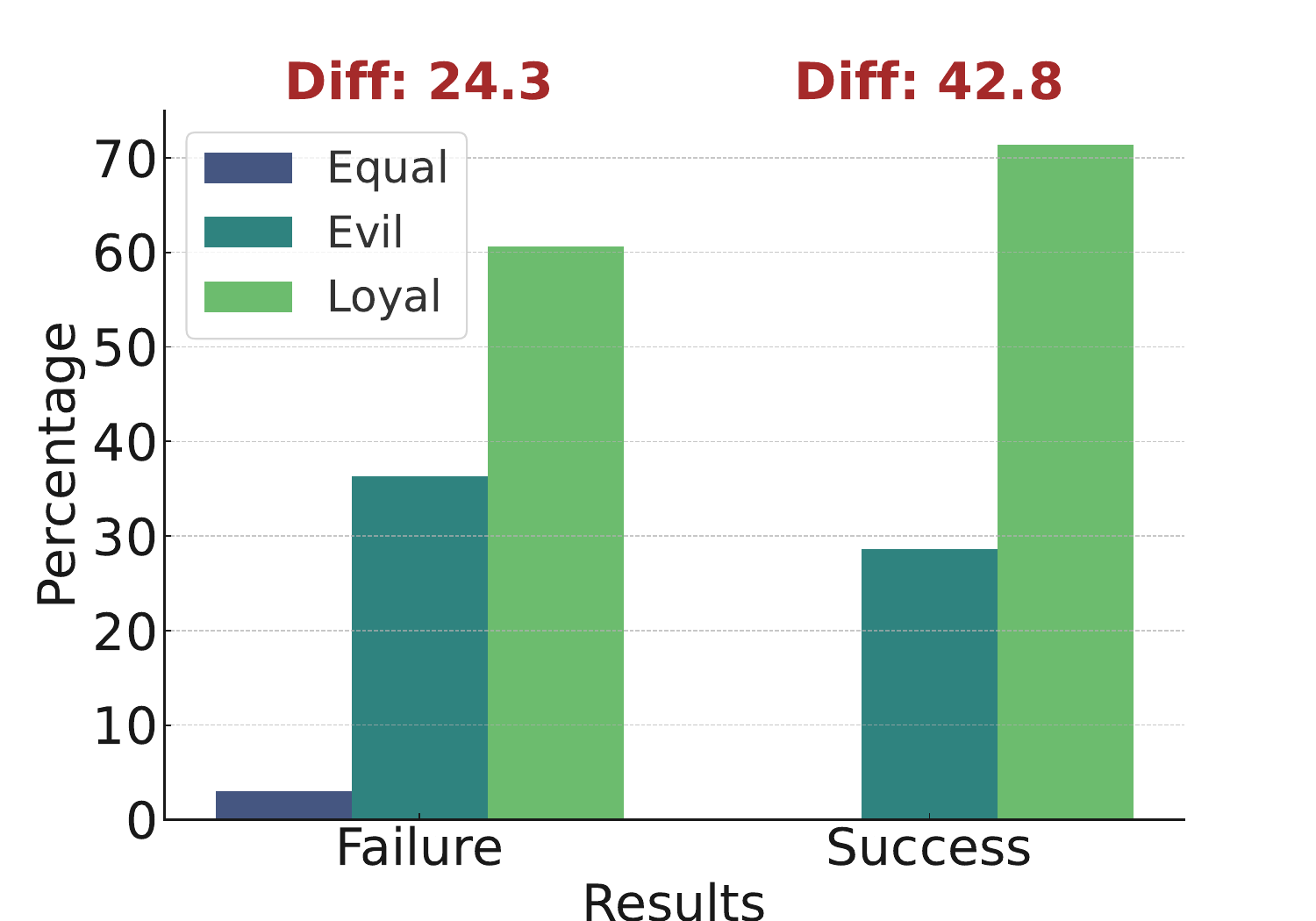}
    \caption{Intention Following $\geq$3 (Game Wise)}
    \label{fig:image2}
  \end{subfigure}
  \hfill
  \begin{subfigure}[b]{0.32\textwidth}
    \centering
    \includegraphics[width=\textwidth]{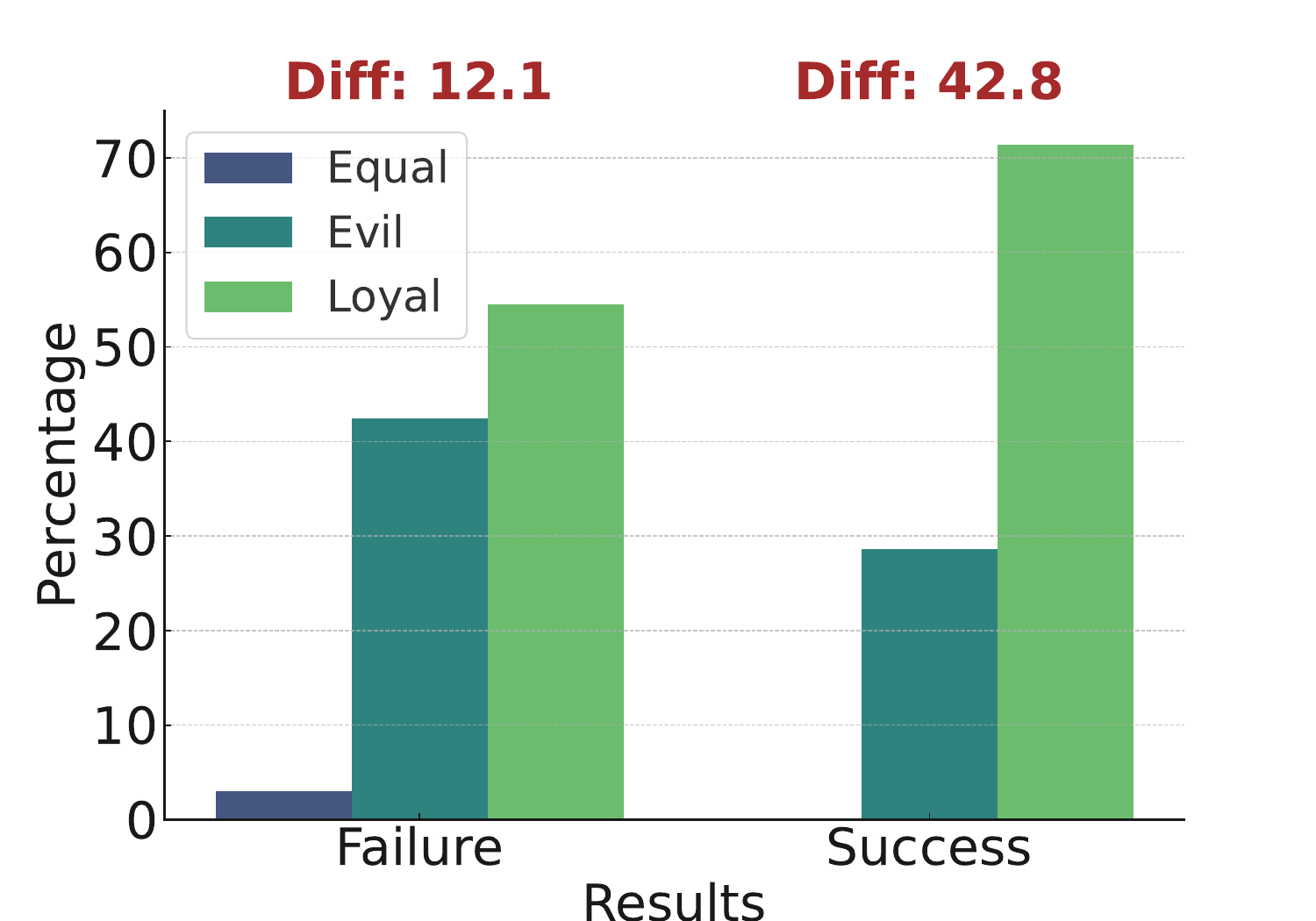}
    \caption{Intention Following $=$3 (Game Wise)}
    \label{fig:image3}
  \end{subfigure}
  \caption{Correlation between Intention Selection/Following and game performance. We present the percentages of games where evil players are equally, better, or worse than loyal players. For example, in games won by loyal players in (a), their performance matches or exceeds that of evil players. We mark the performance differences between evil and loyal players in red, showing a greater gap in successful games/quests compared to failed ones.}
  \label{fig:association}
\end{figure*}

\subsection{Results: Intention Selection and Following}

\paragraph{LLMs show a good understanding of the situation.}
For intention selection, we calculate the number of reasonable intentions over all intentions as accuracy.
The accuracies for GPT-3.5 and GPT-4 are 87.5\% and 88.8\% respectively, indicating that both models effectively understand the situation.
We do not assess whether the selection is optimal due to its subjective nature.
Our focus is solely on whether the models can capture key information and make accurate decisions based on that information.
Instances of unreasonable intentions are primarily attributed to the models forgetting information during lengthy conversations or being influenced by hallucinations from other players.

\paragraph{LLMs can understand intentions abstractly but fall short in planning out intentions.}
For intention following, as shown in Figure \ref{fig:following}, GPT-4 generally outperforms GPT-3.5, in both analytical thinking and articulate speaking.
Specifically in speech, GPT-4's responses tend to be more informative, comprehensive, and detailed.
However, there are instances where GPT-4 deviates from its intended focus, although their discussion still makes sense.
This misalignment with the intentions selected occurs approximately 15\% of the time.
It also appears that adhering to intentions in speaking proves more challenging than in thinking due to stricter evaluation criteria.
This discrepancy suggests that while the model can often conceptualize an appropriate strategy, translating these strategies into clear, actionable responses—particularly in verbal form—remains a complex task.

\begin{figure}
    \centering
    \includegraphics[width=0.48\textwidth]{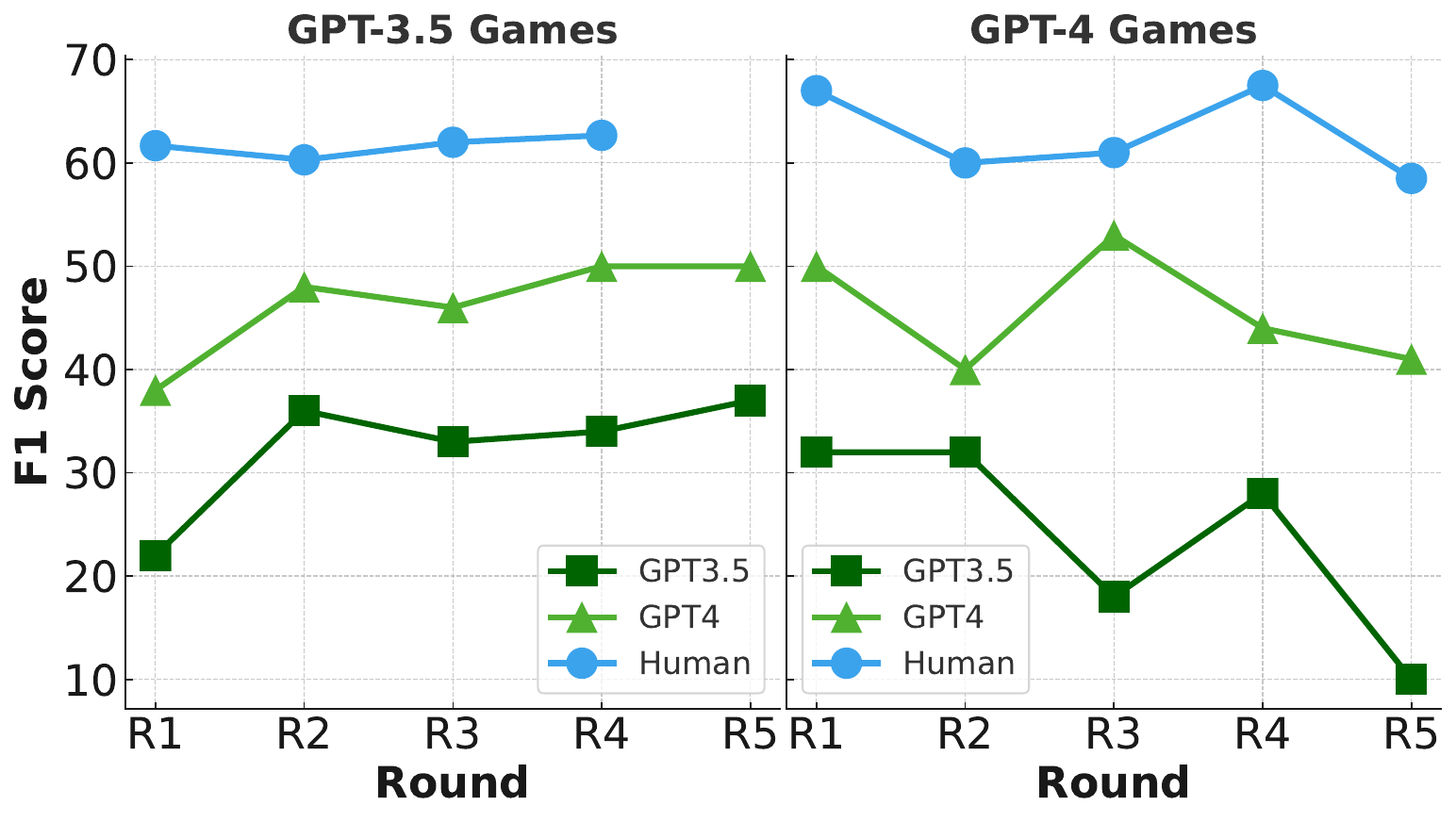}  
    \caption{ToM results over rounds. We provide 200 data points for human results on the GPT-3.5 games, and since usually, games stop at round 4, the results from round 5 are not included.}
    \vspace{-0.3cm}
    \label{fig:over_rounds}
\end{figure}

\paragraph{Intention understanding correlates with game performance.}
To better understand how intention selection and following influence game dynamics, we examine their correlation with both game win rates and quest win rates separately.
For the correlation between intention selection and game performance, we select intentions with a following score greater than two.
For intention following, we consider only reasonable intentions, converting Likert scale responses to binary scores using varying thresholds.
We define $r = \frac{sum(scores)}{n_{teammate}}$ where the score is the binary score result from intention selection/following and $n_{teammate}$ is the number of teammates, two for the evil side and three for the loyal side.
If $r_{evil}>r_{loyal}$, it suggests that the evil players outperformed the loyal players in a given game or round, and vice versa.
We use games from GPT-3.5 for assessment, whose results are shown in Figure~\ref{fig:association} and Figure~\ref{fig:association_supple}.
We observe that loyal players outperform evil players across all quests or games, even in those they ultimately lose.
This discrepancy likely arises from the loyal players' conditions under more constrained information, leading to their occasional losses despite having generally superior performance.
When comparing failed and successful games or quests, the performance gap between loyal and evil players is notably larger in successful ones than in failed ones, indicating that loyal players must significantly outperform evil players to secure a win.
Specifically, in Figure~\ref{fig:association} (c), a threshold of 3 is applied, labeling scores of ``3'' as ``1'' and all others as ``0.''
This threshold can reflect instances of hallucination or incorrect context information.
Hallucinations can significantly influence players, yet incorrect adherence to intentions may occasionally yield positive outcomes.
For example, if a loyal player mistakenly attacks an evil player for an incorrect reason, other players may overlook the mistake and continue to support the action.
Our results suggest a correlation between selecting and following intentions and game performance, especially for loyal players at a disadvantage.
Improved selections and following can enhance overall game performance.

\subsection{Results: Intention Summarization and Guessing}


\paragraph{LLMs can achieve human-level performance in self-awareness but perform much worse than humans in ToM.}
For both summarization and guessing, we report F1 scores.
As shown in Table~\ref{tab:self-awareness}, GPT-3.5 performs below human levels but still achieves respectable scores in self-awareness, while GPT-4 exhibits performance comparable to that of humans.
One key reason for this difference is that ``thinking'' content more explicitly outlines intentions, making it easier for models to extract.
From the ToM results shown in Table~\ref{tab:guessing}, we can see both models struggle significantly compared to self-awareness or compared to human performance.
The experimental setup involves asking GPT-3.5 and GPT-4 to predict each other's intentions, a task differing from self-awareness where a model summarizes its own intentions.
In ToM assessments, models need to understand and predict intentions from responses other than their own, hence the lower performance of GPT-3.5, especially when interpreting the more nuanced and lengthy responses of GPT-4.
This complexity results in GPT-3.5 becoming worse when predicting GPT-4’s responses, while GPT-4 manages similar accuracy in guessing both its own and GPT-3.5’s intentions.
Additionally, human participants generally score better with GPT-4, suggesting that longer contexts in GPT-4's outputs help humans and occasional errors in GPT-3.5’s outputs can sometimes confound human judgment.
 While GPT-4 aligns more closely with human-like processing, there is still considerable potential for improvement in understanding and mimicking human cognitive processes.

\paragraph{Incorporating more context does not necessarily improve the model's ToM ability.}
We also evaluated the impact of information accumulation on the ability to guess intentions, examining how the inclusion of information from previous rounds affects guessing results.
Our analysis shown in Figure~\ref{fig:over_rounds} reveals that for GPT-3.5 game outputs, more information generally leads to improved guessing outcomes.
In contrast, the guessing scores for GPT-4 game outputs fluctuate significantly, with notable declines in GPT-3.5's performance.
This could be attributed to GPT-4 game outputs typically providing more extensive context, which challenges GPT-3.5's capacity to effectively summarize and extract relevant information.
Meanwhile, GPT-4 demonstrates a consistent ability to handle increased data volumes.
Interestingly, human participants maintain similar scores across rounds.
This might stem from humans' ability to directly infer intentions from immediate thinking and speaking cues, without heavily relying on the context. To improve ToM capability, including more context does not necessarily improve the performance but might introduce more complexity to LLMs.

\begin{table}
\centering
\scalebox{0.9}{
\begin{tabular}{ccc}

\toprule
\textbf{Games}&\textbf{GPT-3.5}&\textbf{GPT-4}\\
\midrule
GPT-3.5/4 & $69.49$ & $83.76$\\
Human & $75.21_{\pm0.97}$ & $83.34_{\pm2.72}$ \\
\bottomrule

\end{tabular}
}
\caption{Self-awareness results. The results are GPT-3.5 and GPT-4 summarize their own intentions respectively. We also show the mean and standard deviation of human study results since we include 3 human annotators. We show $F_1$ score in this table.}
\vspace{-0.3cm}
\label{tab:self-awareness}
\end{table}

\section{Related Work}
\label{sec:related_work}

\subsection{Social Intelligence in LLMs}

Artificial social intelligence has long been a significant topic \citep{bainbridge1994artificial,kovač2023socialai,grossmann2023ai,mcdonald2019cognitive,street2024llmtheorymindalignment,li2024socialintelligencedatainfrastructure} as it enables machines to understand and respond to human emotions and social cues, enhancing human-computer interaction and facilitating social science study \citep{ziems2024large,griffin2023large,dubova2022building,gweon2023socially,zhang2024exploringcollaborationmechanismsllm}.
\citet{gweon2023socially} proposed a dataset to evaluate social reasoning and \citet{nematzadeh2018evaluating} crafted a dataset following Sally-Anne experiment to access ToM ability.
Research has demonstrated the emergence and limitations of ToM abilities in LLMs \citep{kosinski2023theory,moghaddam2023boosting,sap2022neural,wu2023coke}.
Although LLMs can understand human intent and choose correct actions in some tasks, their overall performance remains inferior to that of humans.

\subsection{Intentions and Theory of Mind}

ToM, commonly referred to as ``mindreading,'' is traditionally understood as the ability to assign mental states to oneself and others.
This capability is used to interpret and predict actions \citep{van2023theory}.
Despite the emergence of ToM in LLM, some papers argued that LLM learns spurious correlations and ToM ability falls with stress test \citep{shapira2023clever,ullman2023large,sclar2023minding}.
Intentions, as the driving force behind ToM \citep{kennington2022understanding}, is important as the primary purpose of comprehending beliefs and desires is to understand the intentions of others \citep{zelazo2023developing}.
Since social intelligence and ToM can be seen as a society of individually simple ``agents'' \citep{zhuge2023mindstorms}, we focus on intent understanding as an ``agent'' to evaluate social intelligence.

\begin{table}
\centering
\scalebox{0.9}{
\begin{tabular}{ccc}
\toprule
\textbf{Games}&\textbf{GPT-3.5}&\textbf{GPT-4}\\
\midrule
GPT-3.5 & $31.73$ & $24.85$ \\
GPT-4 & $45.66$ & $46.87$ \\
Human & $61.34_{\pm2.75}$ & $65.23_{\pm 5.03}$ \\
\bottomrule
\end{tabular}
}
\caption{ToM results of GPT-3.5 and GPT-4, along with the mean $\pm$ standard deviation of human results. Different from self-awareness, two LLMs also guess each other's intentions. For example, when GPT-4 guesses GPT-3.5 intentions, it achieves a $F_1$ score of 45.66.}
\vspace{-0.3cm}
\label{tab:guessing}
\end{table}

\subsection{Social Game Context}

With the enhancement of LLM performance, more research shifts to social games that necessitate conversation and cooperation \citep{ibraheem2022putting,mansbach2021agent, meta2022human, o2023hoodwinked,park2023generative,zhou2024reallifejustfantasy,wu2024shallteamupexploring}.
\citealt{zhou2023cast} used reinforcement learning to incorporate intent and ToM in DnD games.
Recent works focus on the two most popular board games Avalon~\cite{lan2023llm,shi2023cooperation,wang2023avalons,light2023text} and Werewolf~\cite{xu2023language,xu2023exploring}, exploring social behaviors and strategic playing ability of conversational agents.
Unlike previous work which aimed to improve the game performance, we focus on the in-depth social intelligence evaluation utilizing the Avalon game.
\section{Conclusion}

We introduce a novel framework for evaluating social intelligence within the Avalon context, focusing on intentions.
We examine four critical dimensions: situational awareness, self-regulation, self-awareness, and ToM.
Our findings indicate that models demonstrate a fundamental understanding of situations and self-awareness, but struggle with self-regulation and ToM, particularly when compared to human performance.
Through this systematic approach, our study not only contributes to a deeper understanding of LLMs' social intelligence but also establishes a methodological foundation for future research.
\section*{Limitations}

This study includes several limitations.
First, we evaluate the four essential components of social intelligence but leave several aspects unexplored.
For instance, the capability of a model to modify its intentions post-reflection, \ie, the \textit{Self-Correction}, could also be assessed using our framework.
Future research should consider expanding this approach to include more nuanced facets of social intelligence.
The second limitation lies in our annotation criteria, which are restricted to assessing the reasonableness of intentions to avoid subjectivity.
However, dimensions such as creativity and ingenuity are equally important.
Future studies should aim to measure these attributes to ensure that models not only perform tasks effectively but also demonstrate innovative behaviors indicative of genuine intelligence.
The third limitation stems from the cost.
Half of our evaluations rely on costly human annotation, and the use of GPT-4 incurs significant resource demands due to the extensive context involved.
To make the evaluation process more efficient and cost-effective, we need to develop methods that allow simpler tasks to be managed by smaller models, which introduces additional challenges.
The final limitation is that we did not use an open-source model, as it lacks the capability to effectively engage in social games.

\section*{Ethics Statement}

In order to achieve a smooth human-AI interaction, we need to assess and improve LLMs' intelligence at a social level. Particularly, social intelligence plays an important role in human's daily life and it would be a potential risk if models cannot handle complex social scenarios well. As a first step to building LLMs with human-like social intelligence, our framework provides a testbed to measure the difference between LLMs and humans in social intelligence.

\bibliography{custom}

\begin{thebibliography}{61}
\providecommand{\natexlab}[1]{#1}

\bibitem[{Bainbridge et~al.(1994)Bainbridge, Brent, Carley, Heise, Macy, Markovsky, and Skvoretz}]{bainbridge1994artificial}
William~Sims Bainbridge, Edward~E Brent, Kathleen~M Carley, David~R Heise, Michael~W Macy, Barry Markovsky, and John Skvoretz. 1994.
\newblock Artificial social intelligence.
\newblock \emph{Annual review of sociology}, 20(1):407--436.

\bibitem[{Bandura(1991)}]{bandura1991social}
Albert Bandura. 1991.
\newblock Social cognitive theory of self-regulation.
\newblock \emph{Organizational behavior and human decision processes}, 50(2):248--287.

\bibitem[{Baron-Cohen(1991)}]{baron1991precursors}
Simon Baron-Cohen. 1991.
\newblock Precursors to a theory of mind: Understanding attention in others.
\newblock \emph{Natural theories of mind: Evolution, development and simulation of everyday mindreading}, 1(233-251):1.

\bibitem[{Casanueva et~al.(2020)Casanueva, Tem{\v{c}}inas, Gerz, Henderson, and Vuli{\'c}}]{casanueva2020efficient}
I{\~n}igo Casanueva, Tadas Tem{\v{c}}inas, Daniela Gerz, Matthew Henderson, and Ivan Vuli{\'c}. 2020.
\newblock Efficient intent detection with dual sentence encoders.
\newblock In \emph{Proceedings of the 2nd Workshop on Natural Language Processing for Conversational AI}, pages 38--45.

\bibitem[{Dubova(2022)}]{dubova2022building}
Marina Dubova. 2022.
\newblock Building human-like communicative intelligence: A grounded perspective.
\newblock \emph{Cognitive Systems Research}, 72:63--79.

\bibitem[{Endsley(1995)}]{endsley1995toward}
Mica~R Endsley. 1995.
\newblock Toward a theory of situation awareness in dynamic systems.
\newblock \emph{Human factors}, 37(1):32--64.

\bibitem[{Fitzsimons and Finkel(2010)}]{fitzsimons2010interpersonal}
Gr{\'a}inne~M Fitzsimons and Eli~J Finkel. 2010.
\newblock Interpersonal influences on self-regulation.
\newblock \emph{Current Directions in Psychological Science}, 19(2):101--105.

\bibitem[{Fleiss(1971)}]{fleiss1971measuring}
Joseph~L Fleiss. 1971.
\newblock Measuring nominal scale agreement among many raters.
\newblock \emph{Psychological bulletin}, 76(5):378.

\bibitem[{Gallup et~al.(2003)Gallup, Anderson, and Platek}]{gallup2003self}
GG~Gallup, James~R Anderson, and Steven~M Platek. 2003.
\newblock Self-awareness, social intelligence and schizophrenia.
\newblock \emph{The self in neuroscience and psychiatry}, pages 147--165.

\bibitem[{Gandhi et~al.(2024)Gandhi, Fr{\"a}nken, Gerstenberg, and Goodman}]{gandhi2024understanding}
Kanishk Gandhi, Jan-Philipp Fr{\"a}nken, Tobias Gerstenberg, and Noah Goodman. 2024.
\newblock Understanding social reasoning in language models with language models.
\newblock \emph{Advances in Neural Information Processing Systems}, 36.

\bibitem[{Goodie et~al.(2012)Goodie, Doshi, and Young}]{goodie2012levels}
Adam~S Goodie, Prashant Doshi, and Diana~L Young. 2012.
\newblock Levels of theory-of-mind reasoning in competitive games.
\newblock \emph{Journal of Behavioral Decision Making}, 25(1):95--108.

\bibitem[{Griffin et~al.(2023)Griffin, Kleinberg, Mozes, Mai, Vau, Caldwell, and Mavor-Parker}]{griffin2023large}
Lewis Griffin, Bennett Kleinberg, Maximilian Mozes, Kimberly Mai, Maria Do~Mar Vau, Matthew Caldwell, and Augustine Mavor-Parker. 2023.
\newblock Large language models respond to influence like humans.
\newblock In \emph{Proceedings of the First Workshop on Social Influence in Conversations (SICon 2023)}, pages 15--24.

\bibitem[{Grossmann et~al.(2023)Grossmann, Feinberg, Parker, Christakis, Tetlock, and Cunningham}]{grossmann2023ai}
Igor Grossmann, Matthew Feinberg, Dawn~C Parker, Nicholas~A Christakis, Philip~E Tetlock, and William~A Cunningham. 2023.
\newblock Ai and the transformation of social science research.
\newblock \emph{Science}, 380(6650):1108--1109.

\bibitem[{Gweon et~al.(2023)Gweon, Fan, and Kim}]{gweon2023socially}
Hyowon Gweon, Judith Fan, and Been Kim. 2023.
\newblock Socially intelligent machines that learn from humans and help humans learn.
\newblock \emph{Philosophical Transactions of the Royal Society A}, 381(2251):20220048.

\bibitem[{Huang et~al.(2023)Huang, Wang, Li, Lam, Ren, Yuan, Jiao, Tu, and Lyu}]{huang2023humanity}
Jen-tse Huang, Wenxuan Wang, Eric~John Li, Man~Ho Lam, Shujie Ren, Youliang Yuan, Wenxiang Jiao, Zhaopeng Tu, and Michael Lyu. 2023.
\newblock On the humanity of conversational ai: Evaluating the psychological portrayal of llms.
\newblock In \emph{The Twelfth International Conference on Learning Representations}.

\bibitem[{Ibraheem et~al.(2022)Ibraheem, Zhou, and DeNero}]{ibraheem2022putting}
Samee Ibraheem, Gaoyue Zhou, and John DeNero. 2022.
\newblock Putting the con in context: Identifying deceptive actors in the game of mafia.
\newblock \emph{arXiv preprint arXiv:2207.02253}.

\bibitem[{Kennington(2022)}]{kennington2022understanding}
Casey Kennington. 2022.
\newblock Understanding intention for machine theory of mind: A position paper.
\newblock In \emph{2022 31st IEEE International Conference on Robot and Human Interactive Communication (RO-MAN)}, pages 450--453. IEEE.

\bibitem[{Kosinski(2023)}]{kosinski2023theory}
Michal Kosinski. 2023.
\newblock Theory of mind might have spontaneously emerged in large language models.
\newblock \emph{Preprint at https://arxiv. org/abs/2302.02083}.

\bibitem[{Kovač et~al.(2023)Kovač, Portelas, Dominey, and Oudeyer}]{kovač2023socialai}
Grgur Kovač, Rémy Portelas, Peter~Ford Dominey, and Pierre-Yves Oudeyer. 2023.
\newblock \href {https://arxiv.org/abs/2307.07871} {The socialai school: Insights from developmental psychology towards artificial socio-cultural agents}.
\newblock \emph{Preprint}, arXiv:2307.07871.

\bibitem[{Lan et~al.(2023)Lan, Hu, Wang, Wang, Ye, Zhao, Lim, Xiong, and Wang}]{lan2023llm}
Yihuai Lan, Zhiqiang Hu, Lei Wang, Yang Wang, Deheng Ye, Peilin Zhao, Ee-Peng Lim, Hui Xiong, and Hao Wang. 2023.
\newblock Llm-based agent society investigation: Collaboration and confrontation in avalon gameplay.
\newblock \emph{arXiv preprint arXiv:2310.14985}.

\bibitem[{Le et~al.(2019)Le, Boureau, and Nickel}]{le2019revisiting}
Matthew Le, Y-Lan Boureau, and Maximilian Nickel. 2019.
\newblock Revisiting the evaluation of theory of mind through question answering.
\newblock In \emph{Proceedings of the 2019 Conference on Empirical Methods in Natural Language Processing and the 9th International Joint Conference on Natural Language Processing (EMNLP-IJCNLP)}, pages 5872--5877.

\bibitem[{Li et~al.(2024{\natexlab{a}})Li, Shi, Ziems, and Yang}]{li2024social}
Minzhi Li, Weiyan Shi, Caleb Ziems, and Diyi Yang. 2024{\natexlab{a}}.
\newblock \href {https://arxiv.org/abs/2403.14659} {Social intelligence data infrastructure: Structuring the present and navigating the future}.
\newblock \emph{Preprint}, arXiv:2403.14659.

\bibitem[{Li et~al.(2024{\natexlab{b}})Li, Shi, Ziems, and Yang}]{li2024socialintelligencedatainfrastructure}
Minzhi Li, Weiyan Shi, Caleb Ziems, and Diyi Yang. 2024{\natexlab{b}}.
\newblock \href {https://arxiv.org/abs/2403.14659} {Social intelligence data infrastructure: Structuring the present and navigating the future}.
\newblock \emph{Preprint}, arXiv:2403.14659.

\bibitem[{Liang et~al.(2023)Liang, He, Huang, Wang, Jiao, Wang, Yang, Tu, Shi, and Wang}]{liang2023leveraging}
Tian Liang, Zhiwei He, Jen-tes Huang, Wenxuan Wang, Wenxiang Jiao, Rui Wang, Yujiu Yang, Zhaopeng Tu, Shuming Shi, and Xing Wang. 2023.
\newblock Leveraging word guessing games to assess the intelligence of large language models.
\newblock \emph{arXiv preprint arXiv:2310.20499}.

\bibitem[{Light et~al.(2023{\natexlab{a}})Light, Cai, Shen, and Hu}]{light2023avalonbench}
Jonathan Light, Min Cai, Sheng Shen, and Ziniu Hu. 2023{\natexlab{a}}.
\newblock Avalonbench: Evaluating llms playing the game of avalon.
\newblock In \emph{NeurIPS 2023 Foundation Models for Decision Making Workshop}.

\bibitem[{Light et~al.(2023{\natexlab{b}})Light, Cai, Shen, and Hu}]{light2023text}
Jonathan Light, Min Cai, Sheng Shen, and Ziniu Hu. 2023{\natexlab{b}}.
\newblock From text to tactic: Evaluating llms playing the game of avalon.
\newblock \emph{arXiv preprint arXiv:2310.05036}.

\bibitem[{Malle et~al.(2001)Malle, Moses, and Baldwin}]{malle2001intentions}
Bertram~F Malle, Louis~J Moses, and Dare~A Baldwin. 2001.
\newblock \emph{Intentions and intentionality: Foundations of social cognition}.
\newblock MIT press.

\bibitem[{Mansbach et~al.(2021)Mansbach, Neiterman, and Azaria}]{mansbach2021agent}
Noa Mansbach, Evgeny~Hershkovitch Neiterman, and Amos Azaria. 2021.
\newblock An agent for competing with humans in a deceptive game based on vocal cues.
\newblock In \emph{Interspeech}, pages 4134--4138.

\bibitem[{McDonald and Pearson(2019)}]{mcdonald2019cognitive}
Kelsey~R McDonald and John~M Pearson. 2019.
\newblock Cognitive bots and algorithmic humans: toward a shared understanding of social intelligence.
\newblock \emph{Current Opinion in Behavioral Sciences}, 29:55--62.

\bibitem[{Meta et~al.(2022)Meta, Bakhtin, Brown, Dinan, Farina, Flaherty, Fried, Goff, Gray, Hu et~al.}]{meta2022human}
Fundamental AI Research Diplomacy~Team Meta, Anton Bakhtin, Noam Brown, Emily Dinan, Gabriele Farina, Colin Flaherty, Daniel Fried, Andrew Goff, Jonathan Gray, Hengyuan Hu, et~al. 2022.
\newblock Human-level play in the game of diplomacy by combining language models with strategic reasoning.
\newblock \emph{Science}, 378(6624):1067--1074.

\bibitem[{Moghaddam and Honey(2023)}]{moghaddam2023boosting}
Shima~Rahimi Moghaddam and Christopher~J Honey. 2023.
\newblock Boosting theory-of-mind performance in large language models via prompting.
\newblock \emph{arXiv preprint arXiv:2304.11490}.

\bibitem[{Nematzadeh et~al.(2018)Nematzadeh, Burns, Grant, Gopnik, and Griffiths}]{nematzadeh2018evaluating}
Aida Nematzadeh, Kaylee Burns, Erin Grant, Alison Gopnik, and Thomas~L Griffiths. 2018.
\newblock Evaluating theory of mind in question answering.
\newblock \emph{arXiv preprint arXiv:1808.09352}.

\bibitem[{O'Gara(2023)}]{o2023hoodwinked}
Aidan O'Gara. 2023.
\newblock Hoodwinked: Deception and cooperation in a text-based game for language models.
\newblock \emph{arXiv preprint arXiv:2308.01404}.

\bibitem[{Park et~al.(2023)Park, O'Brien, Cai, Morris, Liang, and Bernstein}]{park2023generative}
Joon~Sung Park, Joseph O'Brien, Carrie~Jun Cai, Meredith~Ringel Morris, Percy Liang, and Michael~S Bernstein. 2023.
\newblock Generative agents: Interactive simulacra of human behavior.
\newblock In \emph{Proceedings of the 36th Annual ACM Symposium on User Interface Software and Technology}, pages 1--22.

\bibitem[{Qiao et~al.(2023)Qiao, Wu, Liang, Li, and Duan}]{qiao2023gameeval}
Dan Qiao, Chenfei Wu, Yaobo Liang, Juntao Li, and Nan Duan. 2023.
\newblock Gameeval: Evaluating llms on conversational games.
\newblock \emph{arXiv preprint arXiv:2308.10032}.

\bibitem[{Sap et~al.(2022)Sap, LeBras, Fried, and Choi}]{sap2022neural}
Maarten Sap, Ronan LeBras, Daniel Fried, and Yejin Choi. 2022.
\newblock Neural theory-of-mind? on the limits of social intelligence in large lms.
\newblock \emph{arXiv preprint arXiv:2210.13312}.

\bibitem[{Sclar et~al.(2023)Sclar, Kumar, West, Suhr, Choi, and Tsvetkov}]{sclar2023minding}
Melanie Sclar, Sachin Kumar, Peter West, Alane Suhr, Yejin Choi, and Yulia Tsvetkov. 2023.
\newblock Minding language models'(lack of) theory of mind: A plug-and-play multi-character belief tracker.
\newblock \emph{arXiv preprint arXiv:2306.00924}.

\bibitem[{Shapira et~al.(2023)Shapira, Levy, Alavi, Zhou, Choi, Goldberg, Sap, and Shwartz}]{shapira2023clever}
Natalie Shapira, Mosh Levy, Seyed~Hossein Alavi, Xuhui Zhou, Yejin Choi, Yoav Goldberg, Maarten Sap, and Vered Shwartz. 2023.
\newblock Clever hans or neural theory of mind? stress testing social reasoning in large language models.
\newblock \emph{arXiv preprint arXiv:2305.14763}.

\bibitem[{Shi et~al.(2023)Shi, Fang, Zheng, Deng, Chen, and Du}]{shi2023cooperation}
Zijing Shi, Meng Fang, Shunfeng Zheng, Shilong Deng, Ling Chen, and Yali Du. 2023.
\newblock Cooperation on the fly: Exploring language agents for ad hoc teamwork in the avalon game.
\newblock \emph{arXiv preprint arXiv:2312.17515}.

\bibitem[{Silvera et~al.(2001)Silvera, Martinussen, and Dahl}]{silvera2001tromso}
David Silvera, Monica Martinussen, and Tove~I Dahl. 2001.
\newblock The troms{\o} social intelligence scale, a self-report measure of social intelligence.
\newblock \emph{Scandinavian journal of psychology}, 42(4):313--319.

\bibitem[{Street(2024)}]{street2024llmtheorymindalignment}
Winnie Street. 2024.
\newblock \href {https://arxiv.org/abs/2405.08154} {Llm theory of mind and alignment: Opportunities and risks}.
\newblock \emph{Preprint}, arXiv:2405.08154.

\bibitem[{Thorndike(1920)}]{thorndike1920intelligence}
Edward~L Thorndike. 1920.
\newblock Intelligence and its uses.
\newblock \emph{Harper's magazine}, 140:227--235.

\bibitem[{Tomasello(2023)}]{tomasello2023having}
Michael Tomasello. 2023.
\newblock Having intentions, understanding intentions, and understanding communicative intentions.
\newblock In \emph{Developing theories of intention}, pages 63--76. Psychology Press.

\bibitem[{Ullman(2023)}]{ullman2023large}
Tomer Ullman. 2023.
\newblock Large language models fail on trivial alterations to theory-of-mind tasks. arxiv.

\bibitem[{van Duijn et~al.(2023)van Duijn, van Dijk, Kouwenhoven, de~Valk, Spruit, and van~der Putten}]{van2023theory}
Max~J van Duijn, Bram van Dijk, Tom Kouwenhoven, Werner de~Valk, Marco~R Spruit, and Peter van~der Putten. 2023.
\newblock Theory of mind in large language models: Examining performance of 11 state-of-the-art models vs. children aged 7-10 on advanced tests.
\newblock \emph{arXiv preprint arXiv:2310.20320}.

\bibitem[{Wang et~al.(2023)Wang, Liu, Zheng, Qi, Chen, Yang, Zhao, Wang, Song, and Huang}]{wang2023avalons}
Shenzhi Wang, Chang Liu, Zilong Zheng, Siyuan Qi, Shuo Chen, Qisen Yang, Andrew Zhao, Chaofei Wang, Shiji Song, and Gao Huang. 2023.
\newblock \href {https://arxiv.org/abs/2310.01320} {Avalon's game of thoughts: Battle against deception through recursive contemplation}.
\newblock \emph{Preprint}, arXiv:2310.01320.

\bibitem[{Wu et~al.(2023)Wu, Chen, Deng, Sabour, and Huang}]{wu2023coke}
Jincenzi Wu, Zhuang Chen, Jiawen Deng, Sahand Sabour, and Minlie Huang. 2023.
\newblock Coke: A cognitive knowledge graph for machine theory of mind.
\newblock \emph{arXiv preprint arXiv:2305.05390}.

\bibitem[{Wu et~al.(2024{\natexlab{a}})Wu, Zhu, Yang, Xu, Fu, Wei, and Fu}]{wu2024enhance}
Shuang Wu, Liwen Zhu, Tao Yang, Shiwei Xu, Qiang Fu, Yang Wei, and Haobo Fu. 2024{\natexlab{a}}.
\newblock \href {https://arxiv.org/abs/2402.02330} {Enhance reasoning for large language models in the game werewolf}.
\newblock \emph{Preprint}, arXiv:2402.02330.

\bibitem[{Wu et~al.(2024{\natexlab{b}})Wu, Peng, Zheng, Liu, Han, Kwon, Onizuka, Tang, and Xiao}]{wu2024shallteamupexploring}
Zengqing Wu, Run Peng, Shuyuan Zheng, Qianying Liu, Xu~Han, Brian~Inhyuk Kwon, Makoto Onizuka, Shaojie Tang, and Chuan Xiao. 2024{\natexlab{b}}.
\newblock \href {https://arxiv.org/abs/2402.12327} {Shall we team up: Exploring spontaneous cooperation of competing llm agents}.
\newblock \emph{Preprint}, arXiv:2402.12327.

\bibitem[{Xu et~al.(2023{\natexlab{a}})Xu, Wang, Li, Luo, Wang, Liu, and Liu}]{xu2023exploring}
Yuzhuang Xu, Shuo Wang, Peng Li, Fuwen Luo, Xiaolong Wang, Weidong Liu, and Yang Liu. 2023{\natexlab{a}}.
\newblock Exploring large language models for communication games: An empirical study on werewolf.
\newblock \emph{arXiv preprint arXiv:2309.04658}.

\bibitem[{Xu et~al.(2023{\natexlab{b}})Xu, Yu, Fang, Wang, and Wu}]{xu2023language}
Zelai Xu, Chao Yu, Fei Fang, Yu~Wang, and Yi~Wu. 2023{\natexlab{b}}.
\newblock Language agents with reinforcement learning for strategic play in the werewolf game.
\newblock \emph{arXiv preprint arXiv:2310.18940}.

\bibitem[{Yott and Poulin-Dubois(2016)}]{yott2016infants}
Jessica Yott and Diane Poulin-Dubois. 2016.
\newblock Are infants’ theory-of-mind abilities well integrated? implicit understanding of intentions, desires, and beliefs.
\newblock \emph{Journal of Cognition and Development}, 17(5):683--698.

\bibitem[{Zelazo et~al.(2023)Zelazo, Astington, and Olson}]{zelazo2023developing}
Philip~David Zelazo, Janet~Wilde Astington, and David~R Olson. 2023.
\newblock \emph{Developing theories of intention: Social understanding and self-control}.
\newblock Psychology Press.

\bibitem[{Zhang et~al.(2024)Zhang, Xu, Zhang, Liu, Hooi, and Deng}]{zhang2024exploringcollaborationmechanismsllm}
Jintian Zhang, Xin Xu, Ningyu Zhang, Ruibo Liu, Bryan Hooi, and Shumin Deng. 2024.
\newblock \href {https://arxiv.org/abs/2310.02124} {Exploring collaboration mechanisms for llm agents: A social psychology view}.
\newblock \emph{Preprint}, arXiv:2310.02124.

\bibitem[{Zhang et~al.(2020)Zhang, Lyu, and Callison-Burch}]{zhang2020intent}
Li~Zhang, Qing Lyu, and Chris Callison-Burch. 2020.
\newblock Intent detection with wikihow.
\newblock In \emph{Proceedings of the 1st Conference of the Asia-Pacific Chapter of the Association for Computational Linguistics and the 10th International Joint Conference on Natural Language Processing}, pages 328--333.

\bibitem[{Zhou et~al.(2023{\natexlab{a}})Zhou, Madaan, Potharaju, Gupta, McKee, Holtzman, Pujara, Ren, Mishra, Nematzadeh et~al.}]{zhou2023far}
Pei Zhou, Aman Madaan, Srividya~Pranavi Potharaju, Aditya Gupta, Kevin~R McKee, Ari Holtzman, Jay Pujara, Xiang Ren, Swaroop Mishra, Aida Nematzadeh, et~al. 2023{\natexlab{a}}.
\newblock How far are large language models from agents with theory-of-mind?
\newblock \emph{arXiv preprint arXiv:2310.03051}.

\bibitem[{Zhou et~al.(2023{\natexlab{b}})Zhou, Zhu, Hu, Pujara, Ren, Callison-Burch, Choi, and Ammanabrolu}]{zhou2023cast}
Pei Zhou, Andrew Zhu, Jennifer Hu, Jay Pujara, Xiang Ren, Chris Callison-Burch, Yejin Choi, and Prithviraj Ammanabrolu. 2023{\natexlab{b}}.
\newblock I cast detect thoughts: Learning to converse and guide with intents and theory-of-mind in dungeons and dragons.
\newblock In \emph{Proceedings of the 61st Annual Meeting of the Association for Computational Linguistics (Volume 1: Long Papers)}, pages 11136--11155.

\bibitem[{Zhou et~al.(2024)Zhou, Su, Eisape, Kim, and Sap}]{zhou2024reallifejustfantasy}
Xuhui Zhou, Zhe Su, Tiwalayo Eisape, Hyunwoo Kim, and Maarten Sap. 2024.
\newblock \href {https://arxiv.org/abs/2403.05020} {Is this the real life? is this just fantasy? the misleading success of simulating social interactions with llms}.
\newblock \emph{Preprint}, arXiv:2403.05020.

\bibitem[{Zhou et~al.(2023{\natexlab{c}})Zhou, Zhu, Mathur, Zhang, Yu, Qi, Morency, Bisk, Fried, Neubig et~al.}]{zhou2023sotopia}
Xuhui Zhou, Hao Zhu, Leena Mathur, Ruohong Zhang, Haofei Yu, Zhengyang Qi, Louis-Philippe Morency, Yonatan Bisk, Daniel Fried, Graham Neubig, et~al. 2023{\natexlab{c}}.
\newblock Sotopia: Interactive evaluation for social intelligence in language agents.
\newblock In \emph{The Twelfth International Conference on Learning Representations}.

\bibitem[{Zhuge et~al.(2023)Zhuge, Liu, Faccio, Ashley, Csord{\'a}s, Gopalakrishnan, Hamdi, Hammoud, Herrmann, Irie et~al.}]{zhuge2023mindstorms}
Mingchen Zhuge, Haozhe Liu, Francesco Faccio, Dylan~R Ashley, R{\'o}bert Csord{\'a}s, Anand Gopalakrishnan, Abdullah Hamdi, Hasan Abed Al~Kader Hammoud, Vincent Herrmann, Kazuki Irie, et~al. 2023.
\newblock Mindstorms in natural language-based societies of mind.
\newblock \emph{arXiv preprint arXiv:2305.17066}.

\bibitem[{Ziems et~al.(2024)Ziems, Held, Shaikh, Chen, Zhang, and Yang}]{ziems2024large}
Caleb Ziems, William Held, Omar Shaikh, Jiaao Chen, Zhehao Zhang, and Diyi Yang. 2024.
\newblock \href {https://arxiv.org/abs/2305.03514} {Can large language models transform computational social science?}
\newblock \emph{Preprint}, arXiv:2305.03514.

\end{thebibliography}

\clearpage
\appendix
\twocolumn
\section{Models and Paramters}
\label{sec:app:models}

We only test GPT-3.5 and GPT-4 in our work as those two are the most common models in the social game study. We also tried open-source model like LLaMA-2 7B and LLaMA-3.1 8B, however, LLaMA-2 could not follow the format instructions properly, as also observed in \citet{wang2023avalons}. LLaMA-3.1 performs better than LLaMA-2 but still fails to follow the format after a few turns of conversations. We set the temperature as 0.8 for both GPT-3.5 and GPT-4. For intention summarization, we also tried sampling five responses and got the aggregated result by majority vote. The result on multiple responses is only 1\% higher than the result on single responses. Therefore, we only generate a single response for self-awareness and ToM tasks. The cost of GPT-3.5 is approximately \$80 and \$25 for GPT-4.

\section{Intentions list}
\subsection{Creation of Intention Set}
\label{sec:app:intentions}

The intentions were created as a combination of domain research and prompting GPT-3.5 to select intention categories from the options listed in Table \ref{tab:intentions} and generate intentions based on the selected categories. For intention category selection we use the prompt shown in Table \ref{tab:intent-category-selection} and for intention generation, we use the prompt shown in Table \ref{tab:intent-generation}. The final set of intentions is listed in Table \ref{tab:intentions}.

\subsection{Selecting Impactful Intention}
\label{app:intention_game}

For selecting a set of impactful intentions, we annotate five complete games played by GPT-3.5. Further, we compute the probability of winning the round given that the intention was selected by a player. To select impactful intentions, we only consider the set of intentions where the computed probability is greater than 0.7 (strong positive association) or it is less than 0.3 (strong negative association) and the intentions selected at least two times for the considered side. We depict the final set of sixteen impactful intentions in red in Table~\ref{tab:intentions}.


\newpage
 

\section{Game Performance}
\label{sec:app:game}

We study the game performance from various levels and perspectives:

\paragraph{Win Rate} 
It is the percentage of games won by the loyal side or the evil side from the total number of games.

\paragraph{Quest Win Rate} 
It is the percentage of quests (game rounds) won by a particular side from the total quests played in all the games. For the loyal side, a successful quest is a quest win, whereas, for the evil team, a failed quest is a quest win.

\paragraph{Quest Engagement Rate} 
It is the percentage of quests (game rounds) where the player gets selected in the quest team from the total quests played in all the games. This metric is calculated for each role separately and averaged across.

\paragraph{Team Selection Accuracy} 
It is the percentage of quests in which the leader correctly selected team members aligned with the objective of their respective side (loyal or evil), out of the total number of quests the leader belonged to that side. Correct team selection for loyal and evil sides is defined as follows: 
\begin{compactitem}
    \item \textbf{Loyal}: The proposed team by the loyal player contains all loyal players.
    \item \textbf{Evil}: The proposed team by the evil player contains at least one evil player.
\end{compactitem}

\paragraph{Failure Vote Rate}
It is the percentage of failure votes from the total votes cast by evil players when they are included in the quest team.

\paragraph{Team Proposal Change Rate}
It is the percentage of game rounds where the leader changes the proposed team after discussion, which reflects the effectiveness of the discussion.

\paragraph{Merlin Assassination Rate}
At the end of the game, Assassin has a chance to identify Merlin. If they identify Merlin, the leader of the loyal side, correctly, then evil players win. This rate is the percentage of games where Assassin correctly identifies Merlin at the end of games over the games where Assassin has a chance to assassin.

We record the entire game and extract intermediate steps, such as voting results, team proposals, and quest outcomes, to calculate game performance.
Results in Table \ref{tab:game_performance} 
indicate that loyal players are at a disadvantage with a notably low game win rate, which corroborates findings from \citet{wang2023avalons}. While the quest win and engagement rates show a balanced performance between loyal and evil players, the evil side demonstrates superior accuracy in team selection. This observation is consistent with their strategic role in the game. The high rate of team proposal changes across all settings suggests a dynamic framework where players consider the discussions of others, an essential aspect of social games. Additionally, the high rate of Merlin assassinations significantly boosts the win rate for evil players. Our objective is not to enhance game performance per se, but to demonstrate that with an intention-guided mechanism, game performance is similar to existing studies, providing a robust testbed for studies in social intelligence.
\begin{table}
\centering
\scalebox{0.80}{
\begin{tabular}{lrrrr}
\toprule
\multirow{2}{*}{Metrics}&\multicolumn{2}{c}{GPT3.5}&\multicolumn{2}{c}{GPT4}\\
\cmidrule{2-3} \cmidrule(lr){4-5}
&Loyal&Evil&Loyal&Evil\\
\midrule
Win rate & 18.0&82.0&40.0&60.0\\
Quest win rate & 41.4&58.6&54.5&45.5\\
Quest engagement rate & 53.6&43.5&59.1&38.6\\
Team selection accuracy & 32.1&80.0&71.4&100.0\\
Failure vote rate & -&79.9&-&86.7\\
Team proposal change rate & 76.2&74.1&50.00&62.5\\
Merlin assassination rate & -&41.7&-&33.3\\
\bottomrule

\end{tabular}
}
\caption{Game performance results shown in percentage. For the failure vote rate and Merlin assassination rate, there is no result for Loyal players as they are not allowed to vote failure or assassinate Merlin.}
\label{tab:game_performance}
\end{table}
\section{Structured Context Prompts}
\label{sec:appendix:structured_con}

For intention summarization and intention guessing, we use structured context prompts. The detailed prompt structures are shown in Table \ref{tab:structured_context_intent_sum} and Table \ref{tab:structured_context_intent_guess}.
\section{Game Pipeline Prompts}
\label{sec:app:game_prompts}
\begin{figure*}[h]
    \centering
    \includegraphics[width=\linewidth]{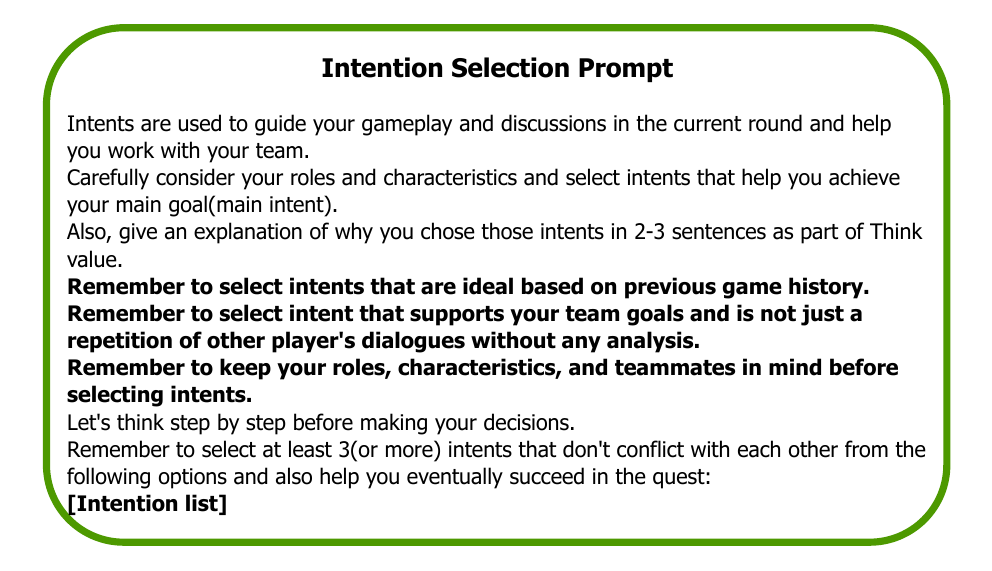}
    \caption{Intention selection prompt.}
    \label{fig:selection_prompt}
\end{figure*}

\begin{figure*}
    \centering
    \includegraphics[width=\linewidth]{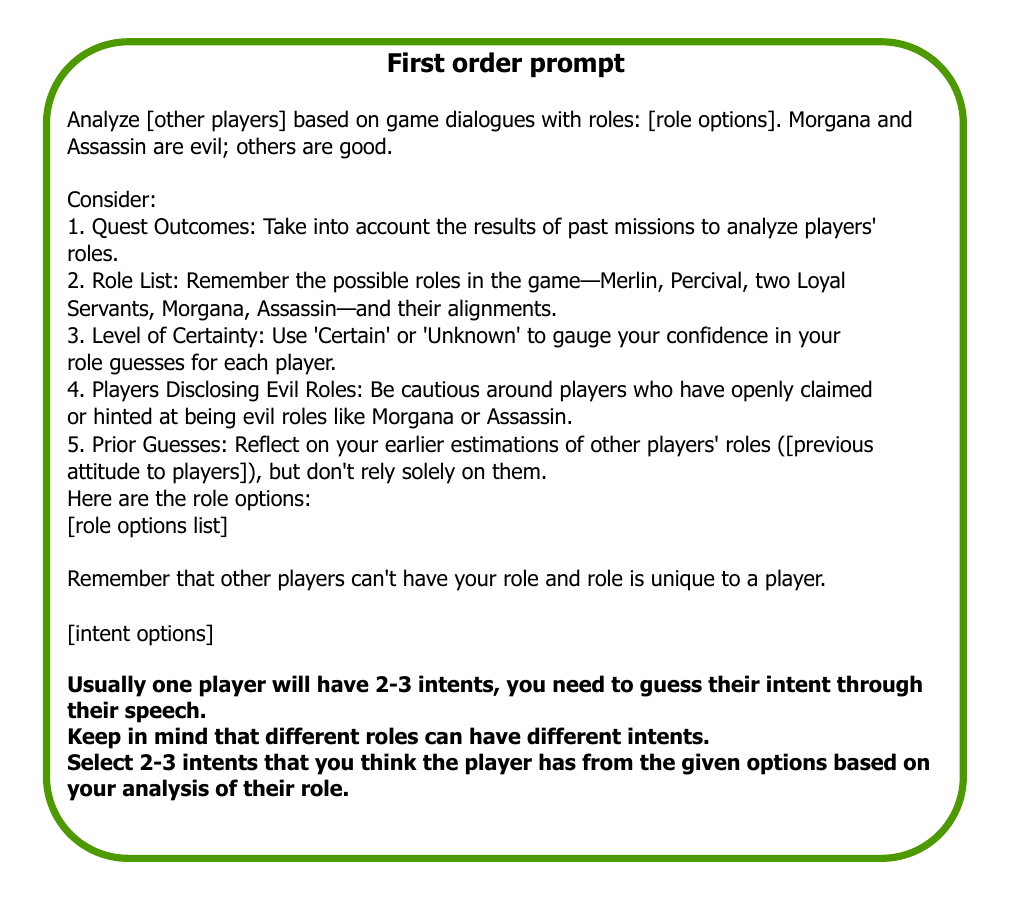}
    \caption{First order prompt.}
    \label{fig:first_prompt}
\end{figure*}

\begin{figure*}
    \centering
    \includegraphics[width=\linewidth]{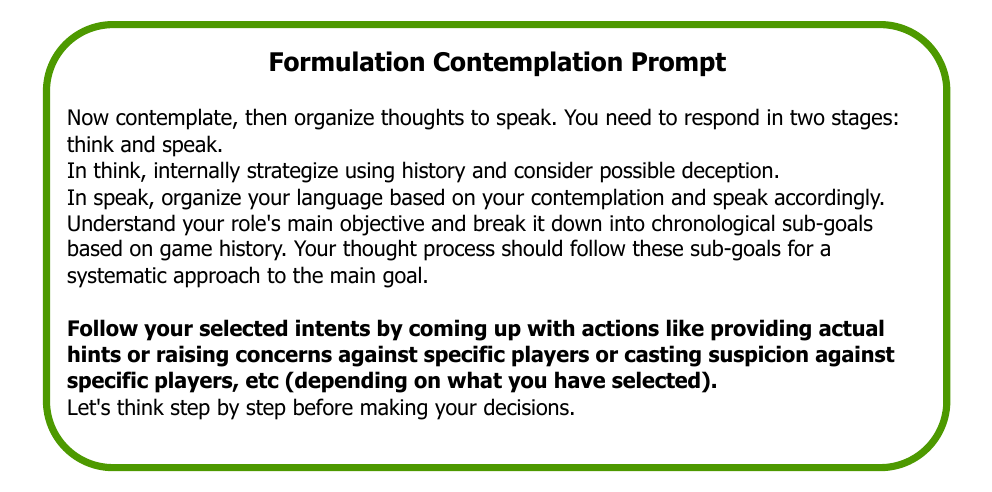}
    \caption{Formulation contemplation prompt.}
    \label{fig:formulation_prompt}
\end{figure*}

\begin{figure*}
    \centering
    \includegraphics[width=\linewidth]{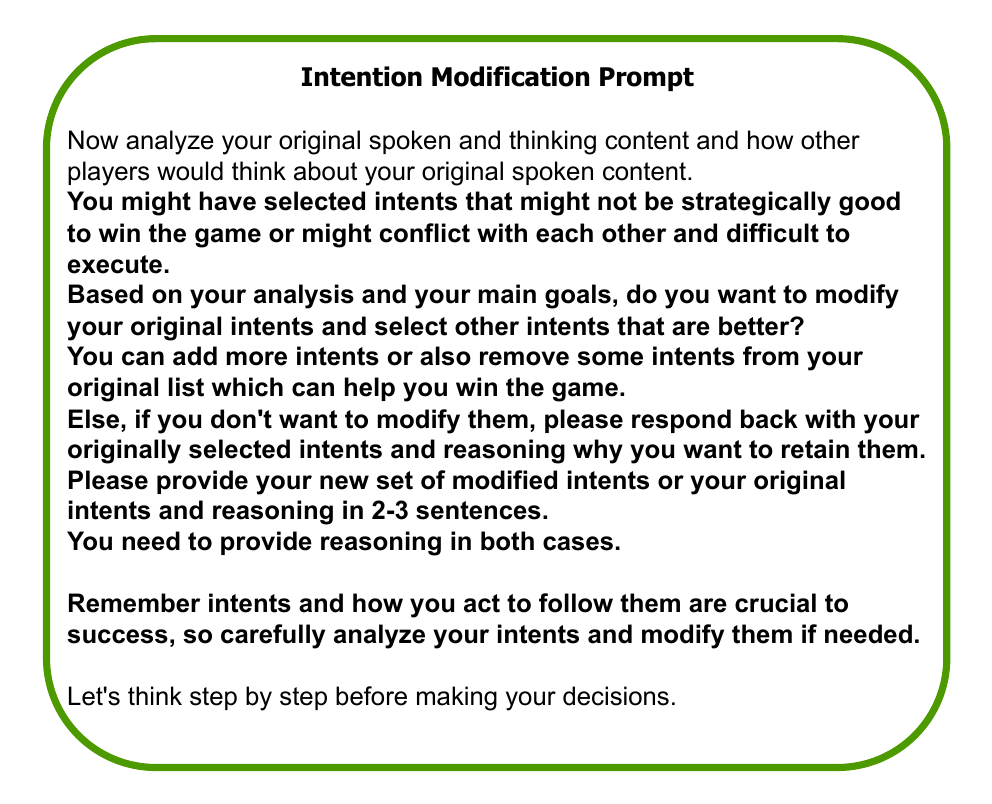}
    \caption{Intention modification prompt.}
    \label{fig:modification_prompt}
\end{figure*}

\begin{figure*}
    \centering
    \includegraphics[width=\linewidth]{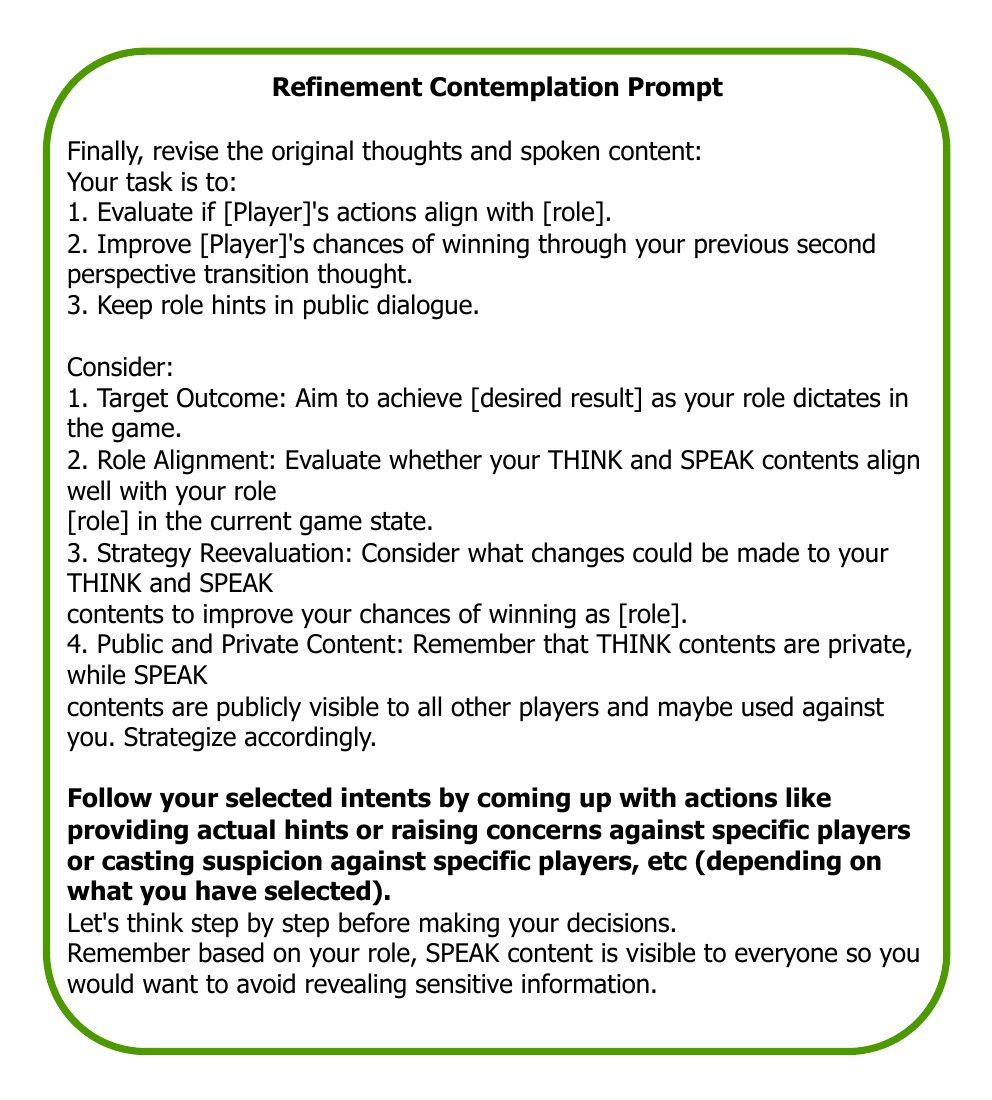}
    \caption{Refinement contemplation prompt.}
    \label{fig:refinement_prompt}
\end{figure*}

\begin{figure*}
    \centering
    \includegraphics[width=\linewidth]{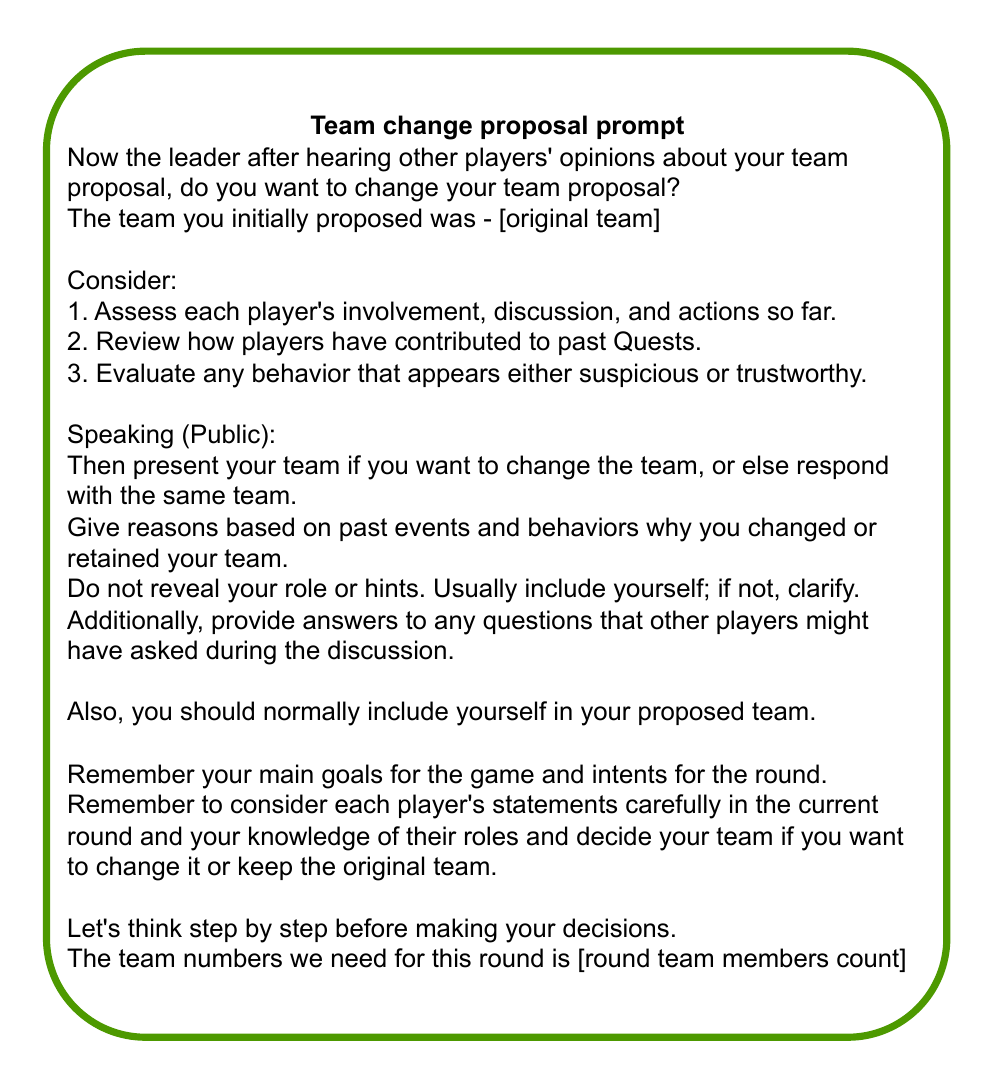}
    \caption{Team proposal change prompt.}
    \label{fig:change_prompt}
\end{figure*}

\begin{figure*}
    \centering
    \includegraphics[width=\linewidth]{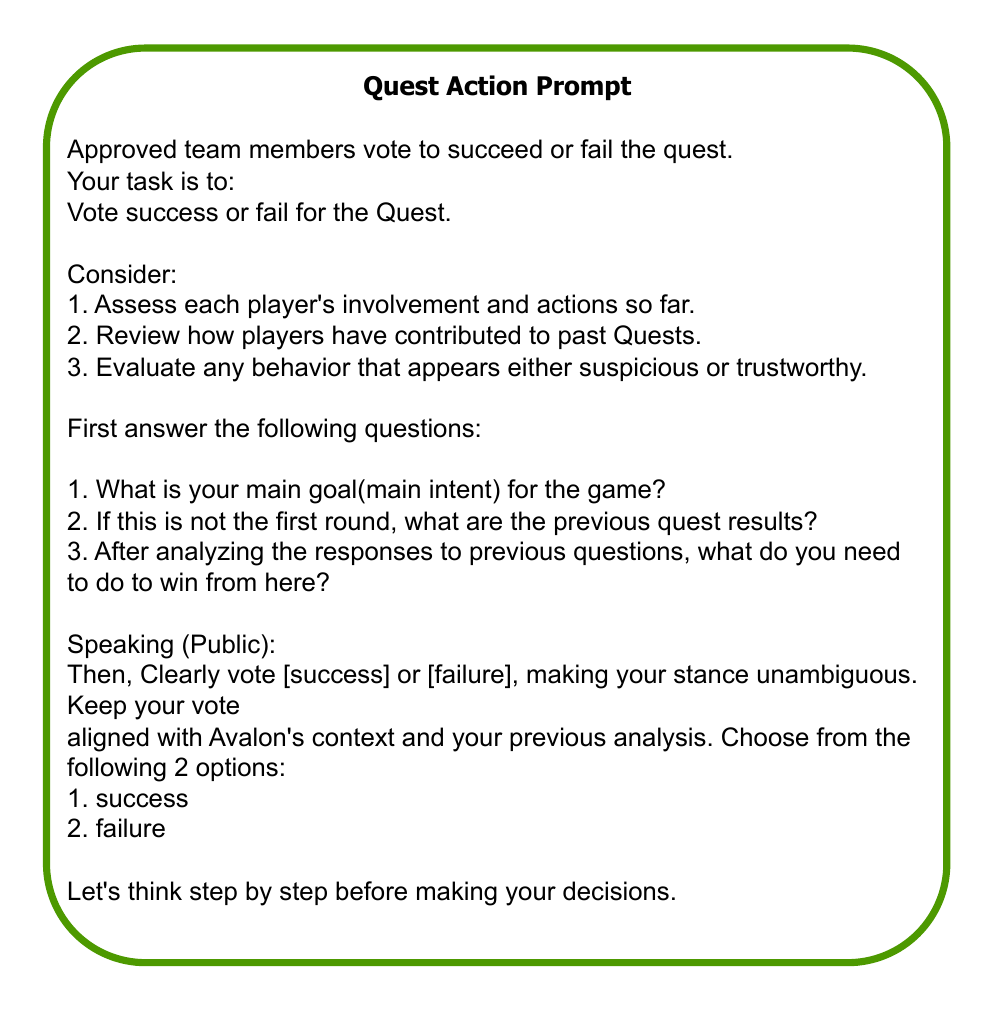}
    \caption{Quest action prompt.}
    \label{fig:quest_prompt}
\end{figure*}

Since we mostly refer to the prompt by \citet{wang2023avalons}, we only present prompts different from theirs to reduce the redundancy. The prompts we use are shown in Figure \ref{fig:selection_prompt} to Figure \ref{fig:quest_prompt}. For the first order, formulation contemplation, and refinement contemplation, we add an intention mechanism based on original prompts. We also add several questions in the quest action prompt to help evil players better choose action.

\section{Human Annotation Details}
\label{sec:app:anno}
\begin{figure*}
    \centering
    \includegraphics[width=\linewidth]{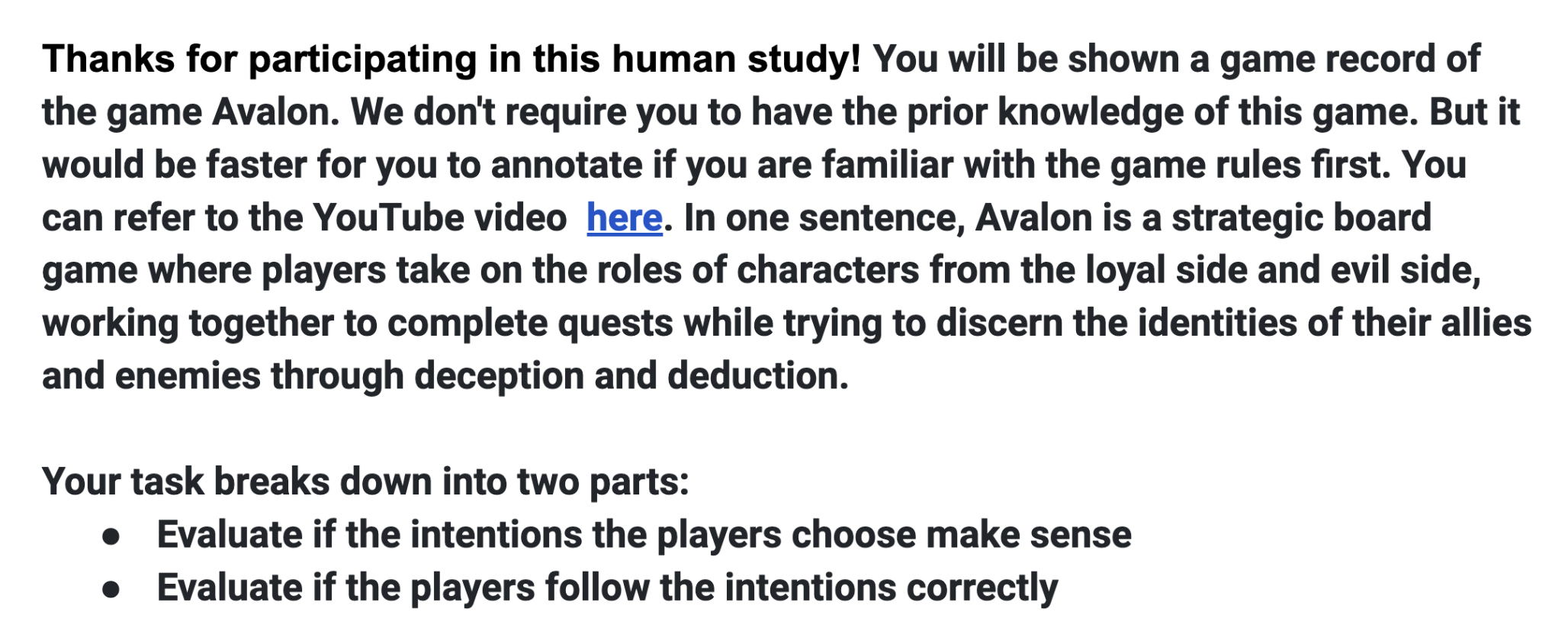}
    \caption{Introduction for the human annotations and human study.}
    \label{fig:human_intro}
\end{figure*}

\begin{figure*}
    \centering
    \includegraphics[width=\linewidth]{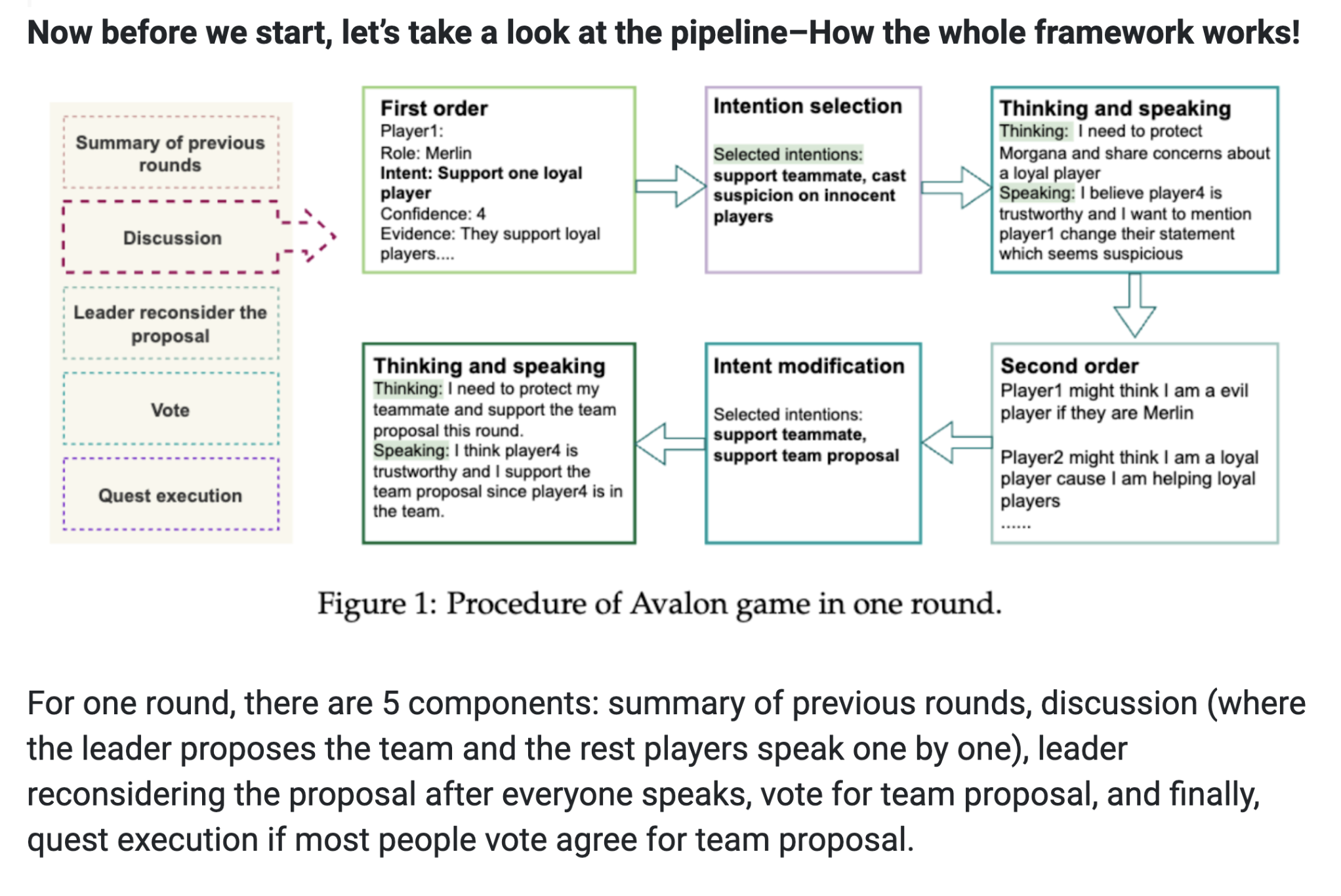}
    \caption{Explanation of framework.}
    \label{fig:human_framework}
\end{figure*}

\begin{figure*}
    \centering
    \includegraphics[width=\linewidth]{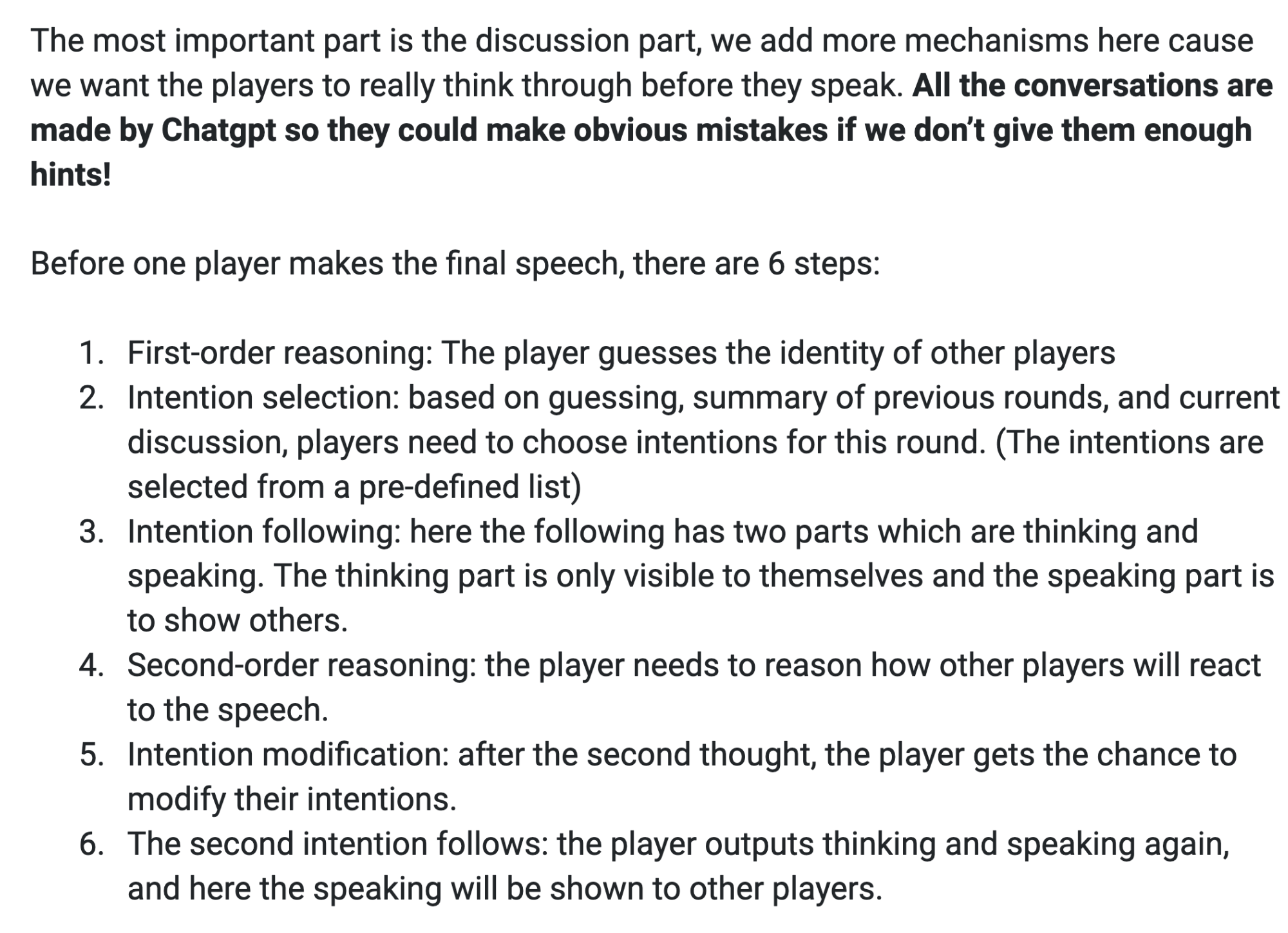}
    \caption{Explanation of 6 steps in the game.}
    \label{fig:human_6 steps}
\end{figure*}

\begin{figure*}
    \centering
    \includegraphics[width=\linewidth]{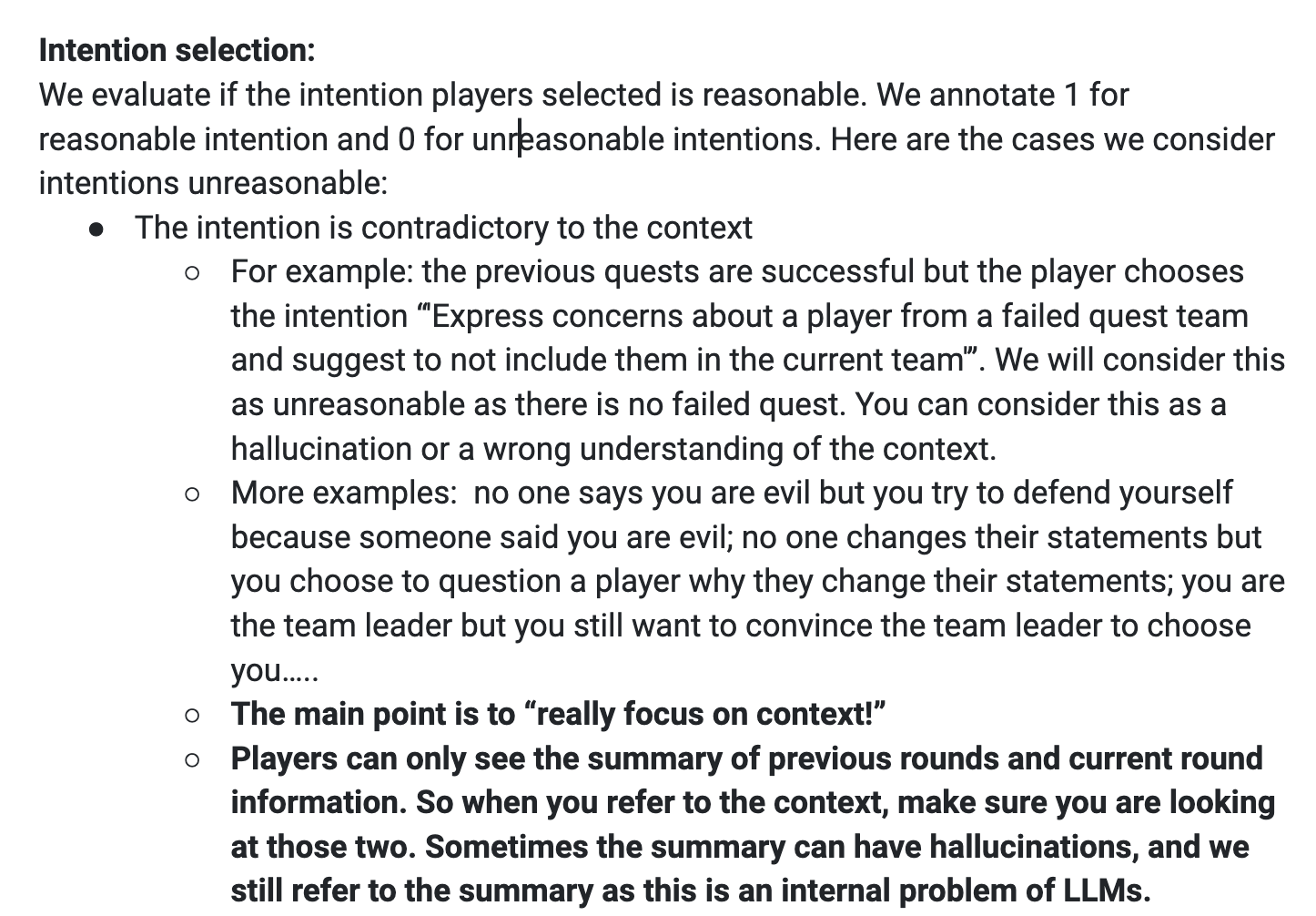}
    \caption{Intent selection annotation instruction.}
    \label{fig:human_is_1}
\end{figure*}

\begin{figure*}
    \centering
    \includegraphics[width=\linewidth]{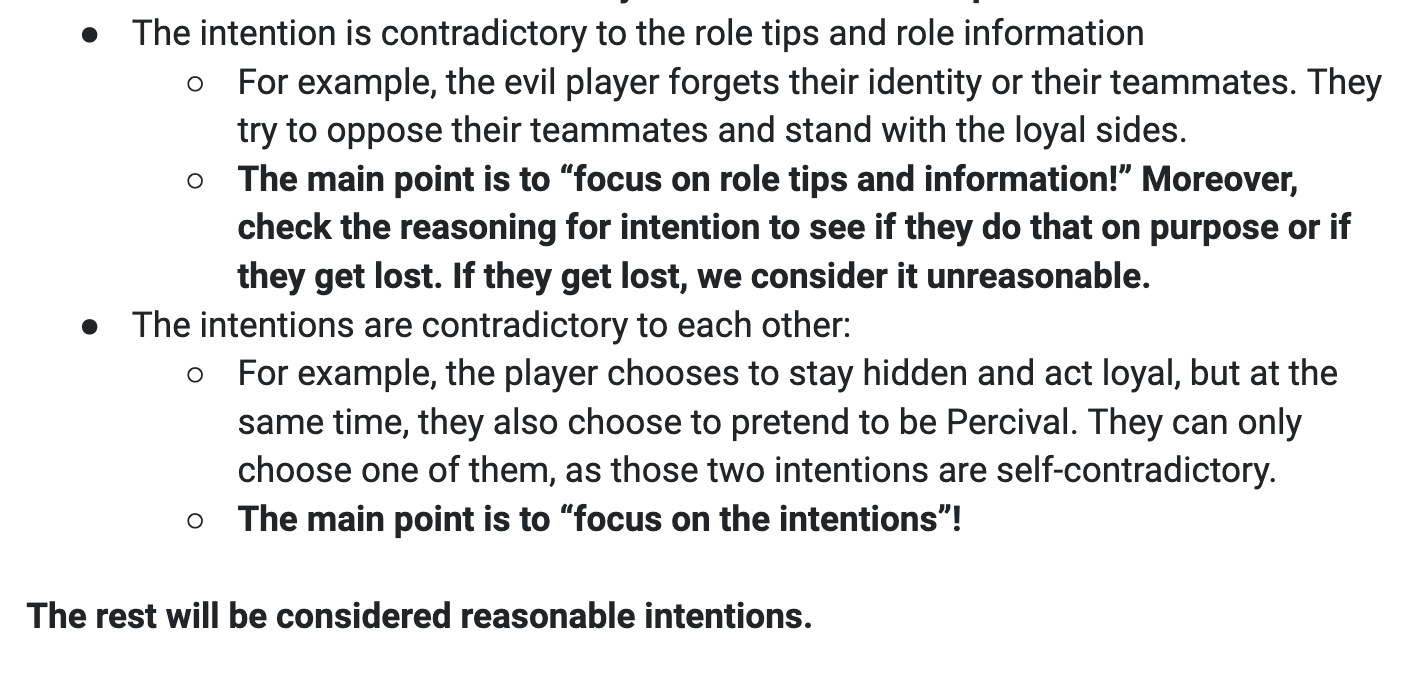}
    \caption{Intent selection annotation instruction.}
    \label{fig:human_is_2}
\end{figure*}

\begin{figure*}
    \centering
    \includegraphics[width=\linewidth]{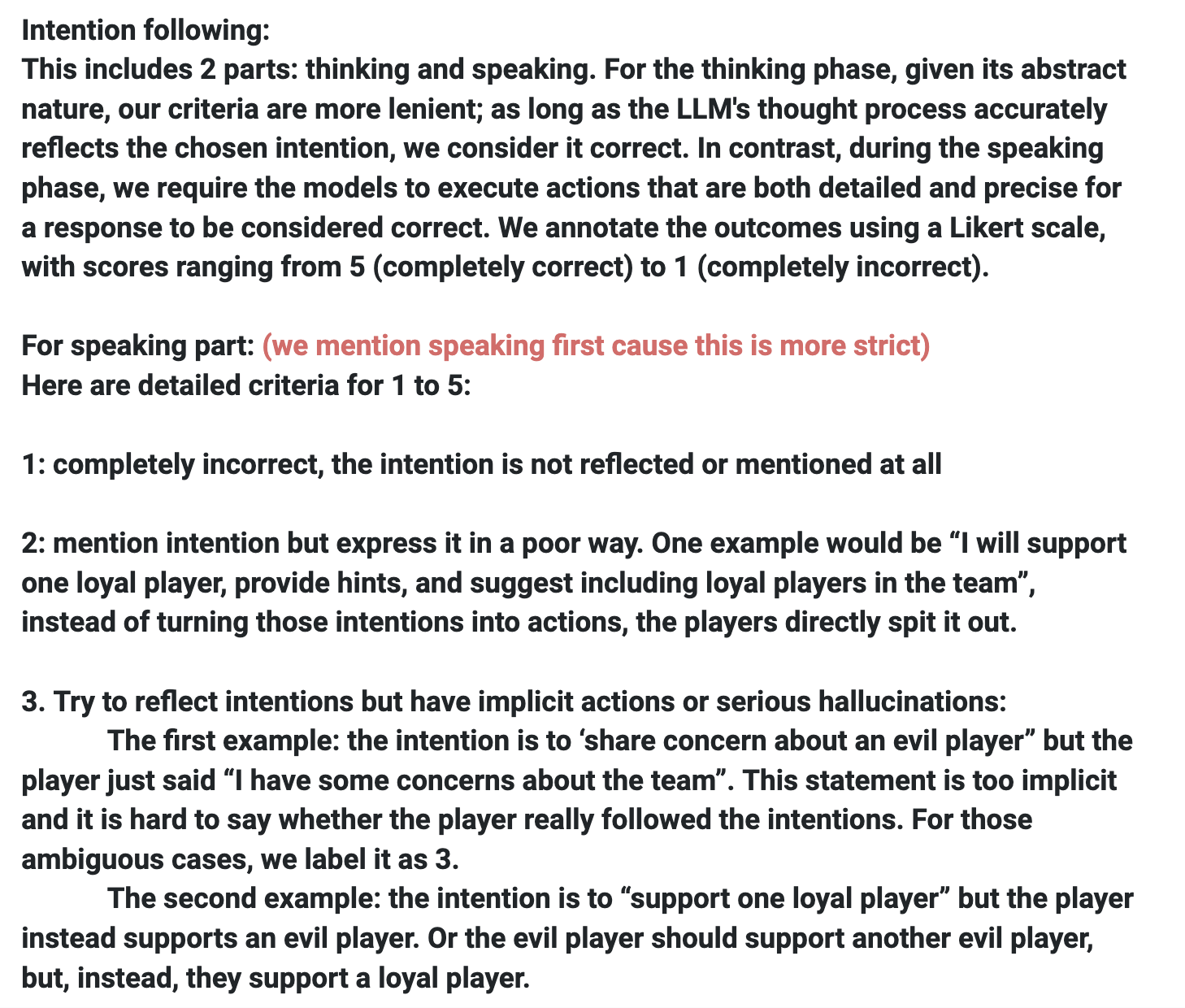}
    \caption{Intent following annotation instruction.}
    \label{fig:human_if_1}
\end{figure*}

\begin{figure*}
    \centering
    \includegraphics[width=\linewidth]{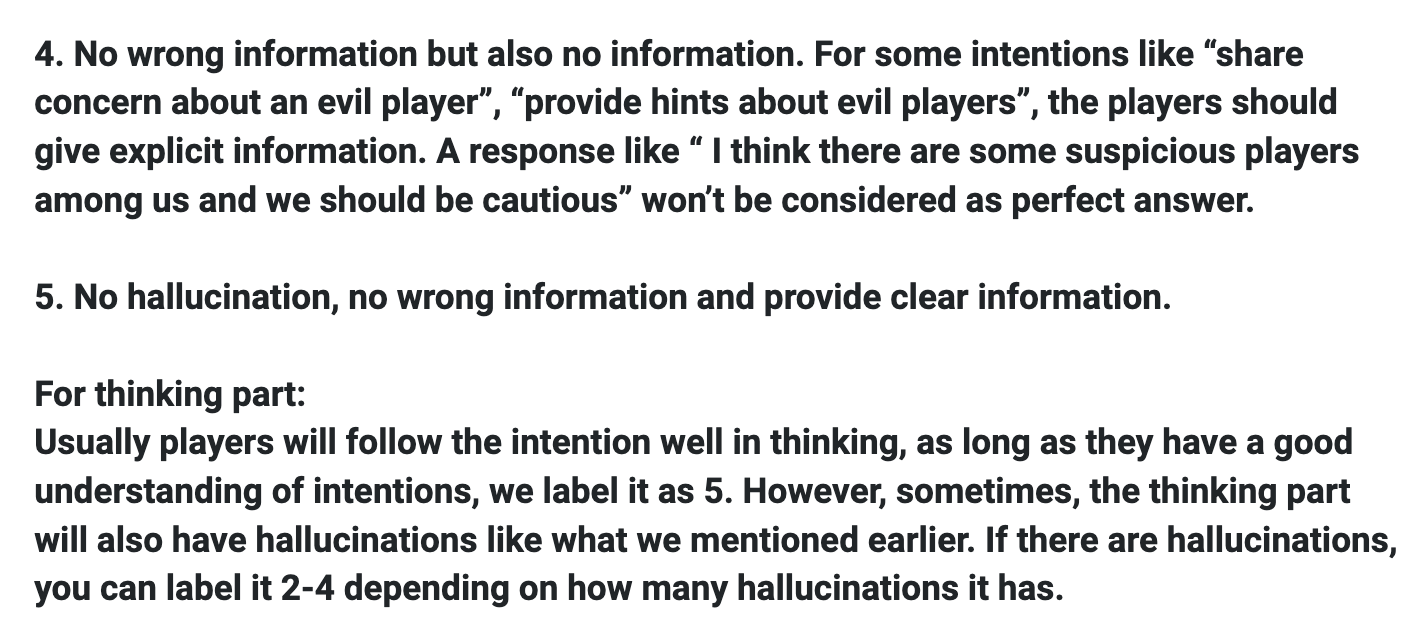}
    \caption{Intent following annotation instruction.}
    \label{fig:human_if_2}
\end{figure*}

\begin{figure*}
    \centering
    \includegraphics[width=\linewidth]{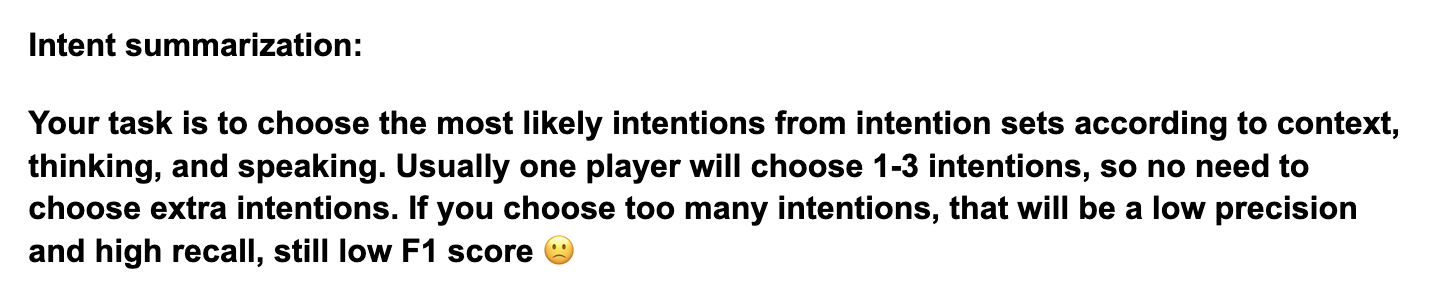}
    \caption{Intent summarization study instruction.}
    \label{fig:human_is}
\end{figure*}

\begin{figure*}
    \centering
    \includegraphics[width=\linewidth]{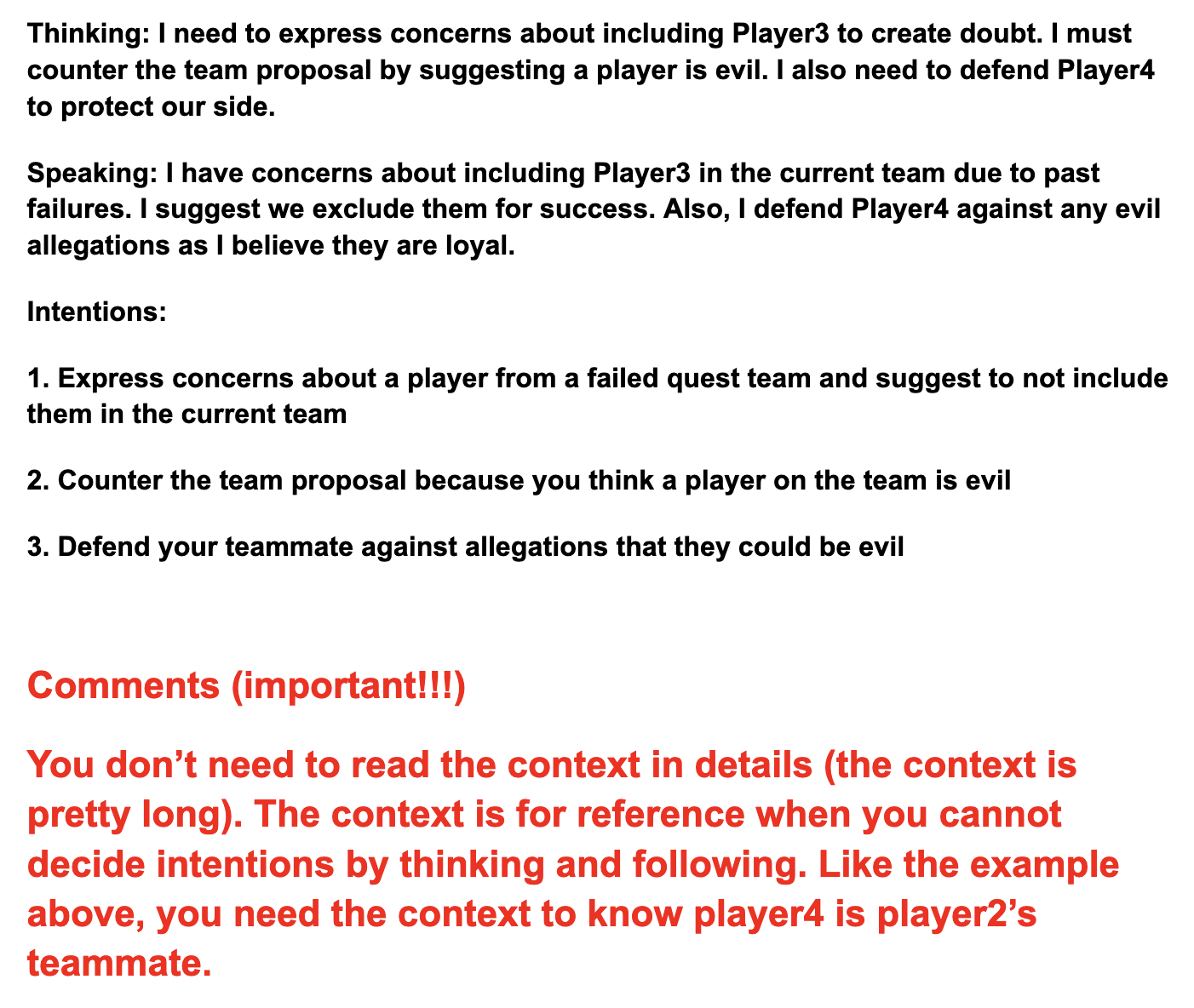}
    \caption{Intent summarization study instruction.}
    \label{fig:human_is_example}
\end{figure*}

\begin{figure*}
    \centering
    \includegraphics[width=\linewidth]{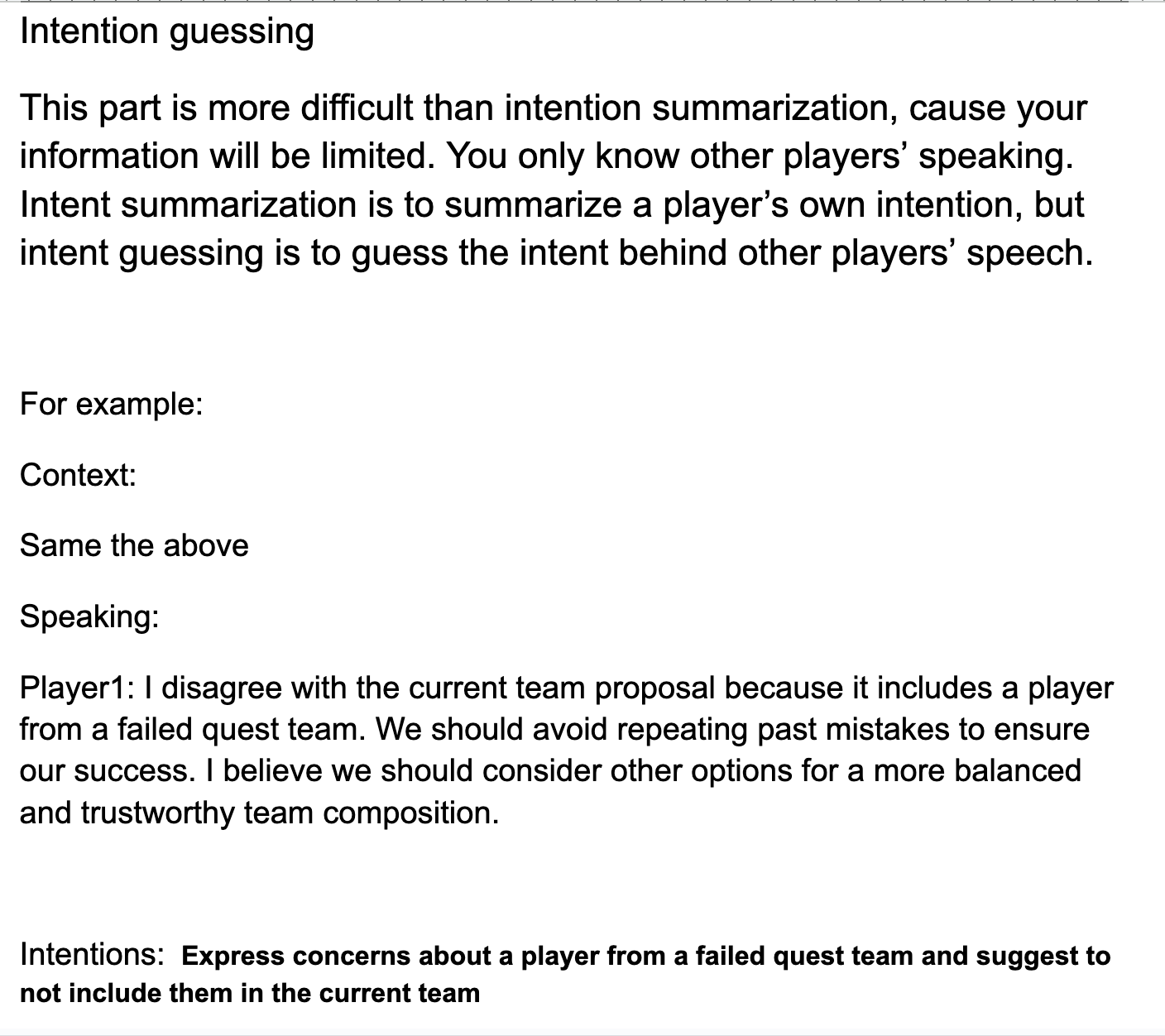}
    \caption{Intent guessing study instruction.}
    \label{fig:human_ig}
\end{figure*}

We show all the instructions we use in this section, from Figure \ref{fig:human_intro} to Figure \ref{fig:human_ig}.
For intention summarization and guessing, we also illustrate the whole intention lists shown in Table \ref{tab:intentions}.

We experimented with several grouping methods to calculate agreement scores. When using the grouping of 1 $\sim$ 4 and 5, the Fleiss' Kappa scores for "thinking" and "speaking" were 0.448 and 0.420, respectively. These scores are relatively lower, likely due to differing interpretations of informativeness among human annotators. For the grouping of 1 $\sim$ 2 and 3 $\sim$ 5, the agreement was low, as scores of 1 $\sim$ 2 accounted for only a small portion of the entire dataset. Since Fleiss' Kappa is highly sensitive to disagreements within the 1 $\sim$ 2 group, even minor discrepancies resulted in an overall lower score.  
\begin{table*}[t!]
\resizebox{1.0\linewidth}{!}{
\begin{tabular}{l}
\toprule
\textbf{Intention Category Selection Prompt}\\
\midrule
\makecell[l]{
Intentions are used to guide your game playing and they should align with your game Goal(main intent) and your role.\\
Your selected intentions should support your main intent.\\
Now select multiple intention categories that are ideal for this round depending on your role and strategy from the\\ options listed below.
Also give explanation of why you chose those categories in 2-3 sentences as part of Think value.\\
Your explanation won't be shown to others.\\
\\
Let's think step by step before making your decisions.\\
Remember to select at least 3 or more intentions from the given options only. intention category options are:\\
$[$intentioncategory options$]$\\
}\\
\bottomrule
\end{tabular}
}
\caption{Intentions category selection prompt.}
\label{tab:intent-category-selection}
\end{table*}

\begin{table*}[t!]
\resizebox{1.0\linewidth}{!}{
\begin{tabular}{l}
\toprule
\textbf{Intention Generation Prompt}\\
\midrule
\makecell[l]{
First answer the following questions:\\
1. What is your main goal(main intent) for the game?\\
2. What have you observed from discussions in this round?\\
3. What should be the ideal team composition for your side to win?\\
4. After analyzing the responses to previous questions, what do you need to do to win from here?\\
\\
Consider answers to your previous questions and your roles and characteristics and generate intentions that help you achieve\\ your main goal (main intention).\\
Please do not repeat intentions and ensure they are different from each other.\\
Each intention should be 10-15 words long.\\
\\
Let's think step by step before making your decisions.\\
Remember to generate at least 3 intentions, with a minimum of one intention for each intention category you have selected below:\\
$[$intentioncategories selected by player$]$\\
}\\
\bottomrule
\end{tabular}
}
\caption{Intentions generation prompt.}
\label{tab:intent-generation}
\end{table*}


\begin{figure*}[h]
  \centering
  \begin{subfigure}[b]{0.32\textwidth}
    \centering
    \includegraphics[width=\textwidth]{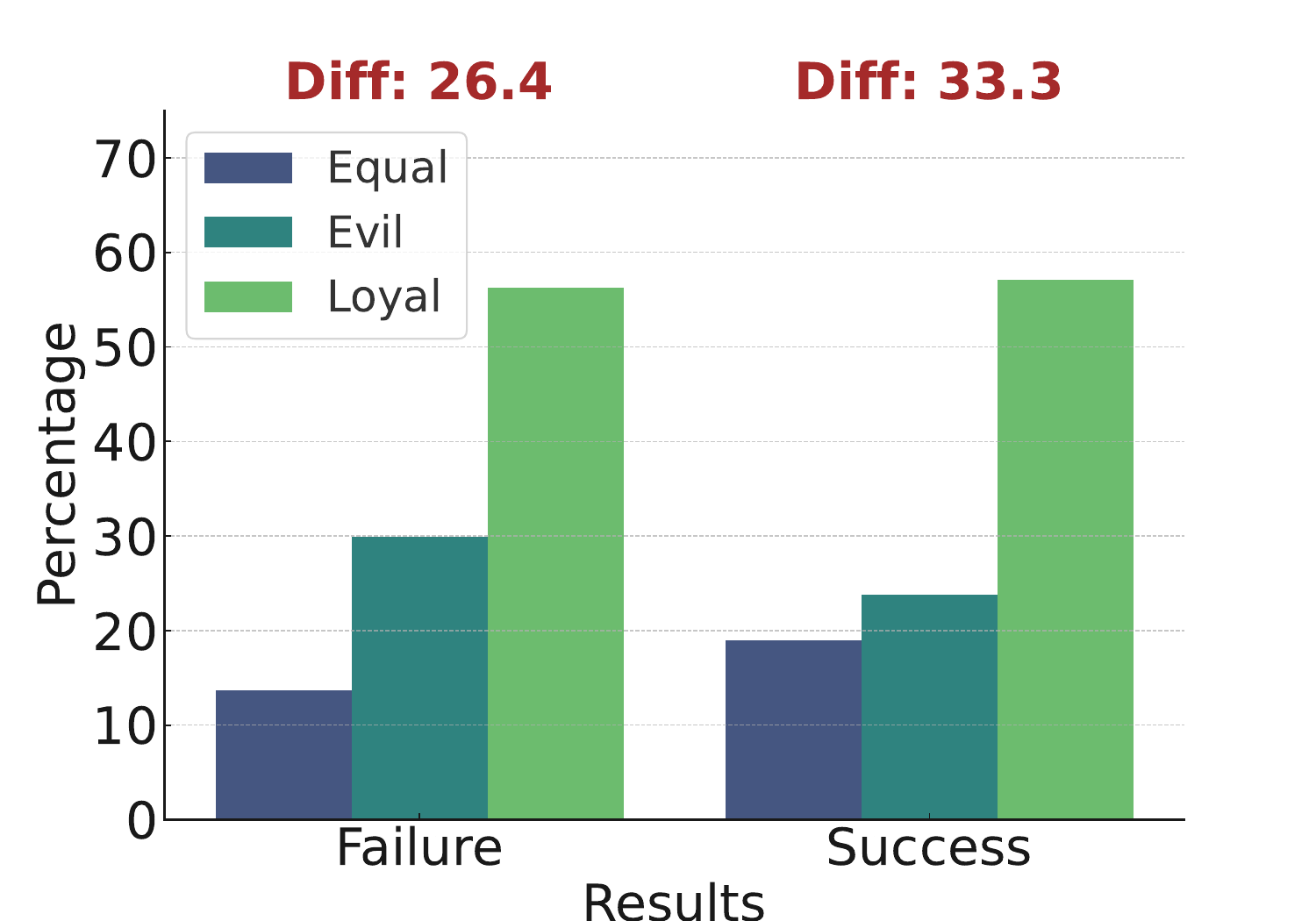}
    \caption{Intention Selection (Round Wise)}
    \label{fig:image4}
  \end{subfigure}
  \hfill
  \begin{subfigure}[b]{0.32\textwidth}
    \centering
    \includegraphics[width=\textwidth]{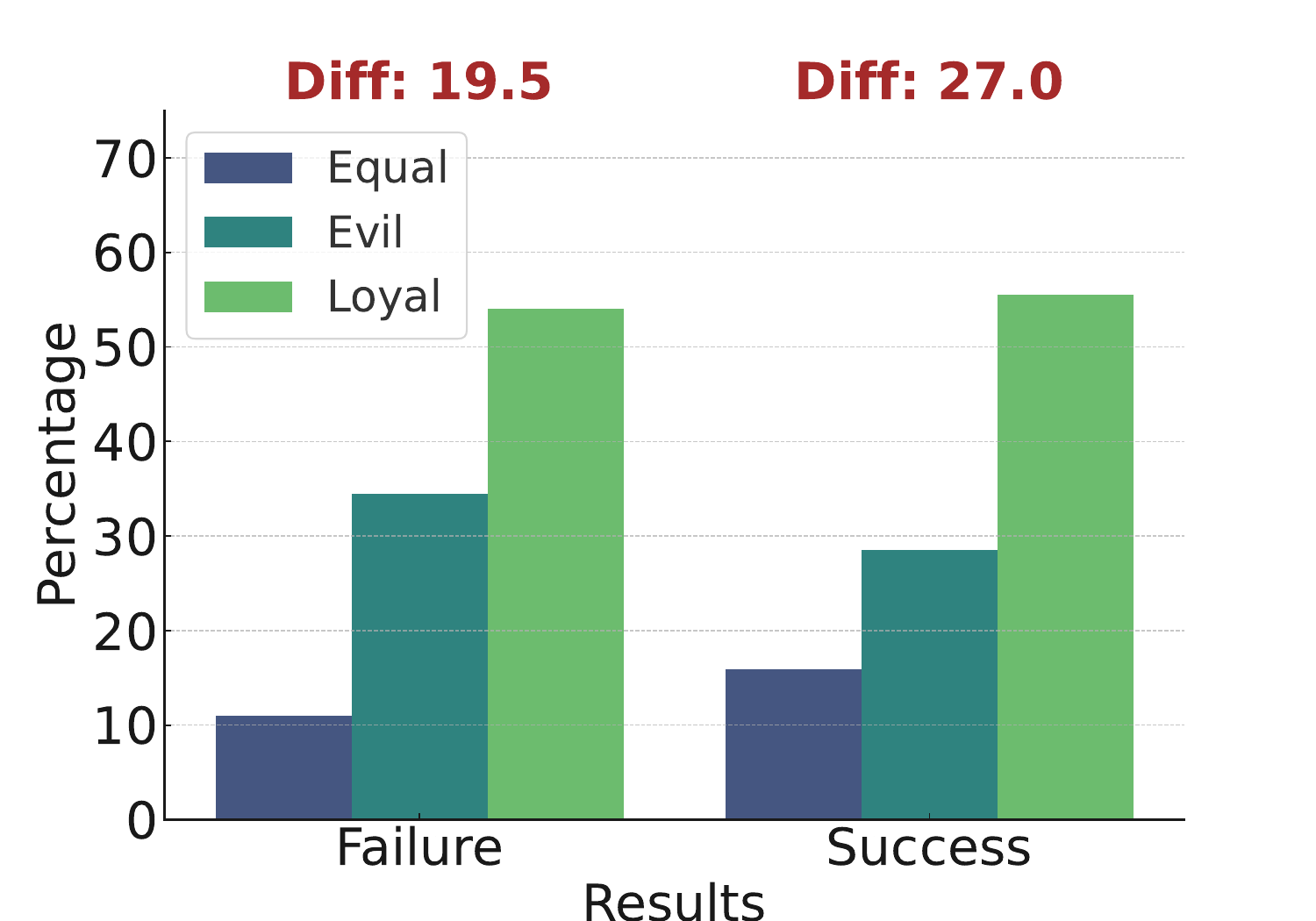}
    \caption{Intention Following $\geq$3 (Round Wise)}
    \label{fig:image5}
  \end{subfigure}
  \hfill
  \begin{subfigure}[b]{0.32\textwidth}
    \centering
    \includegraphics[width=\textwidth]{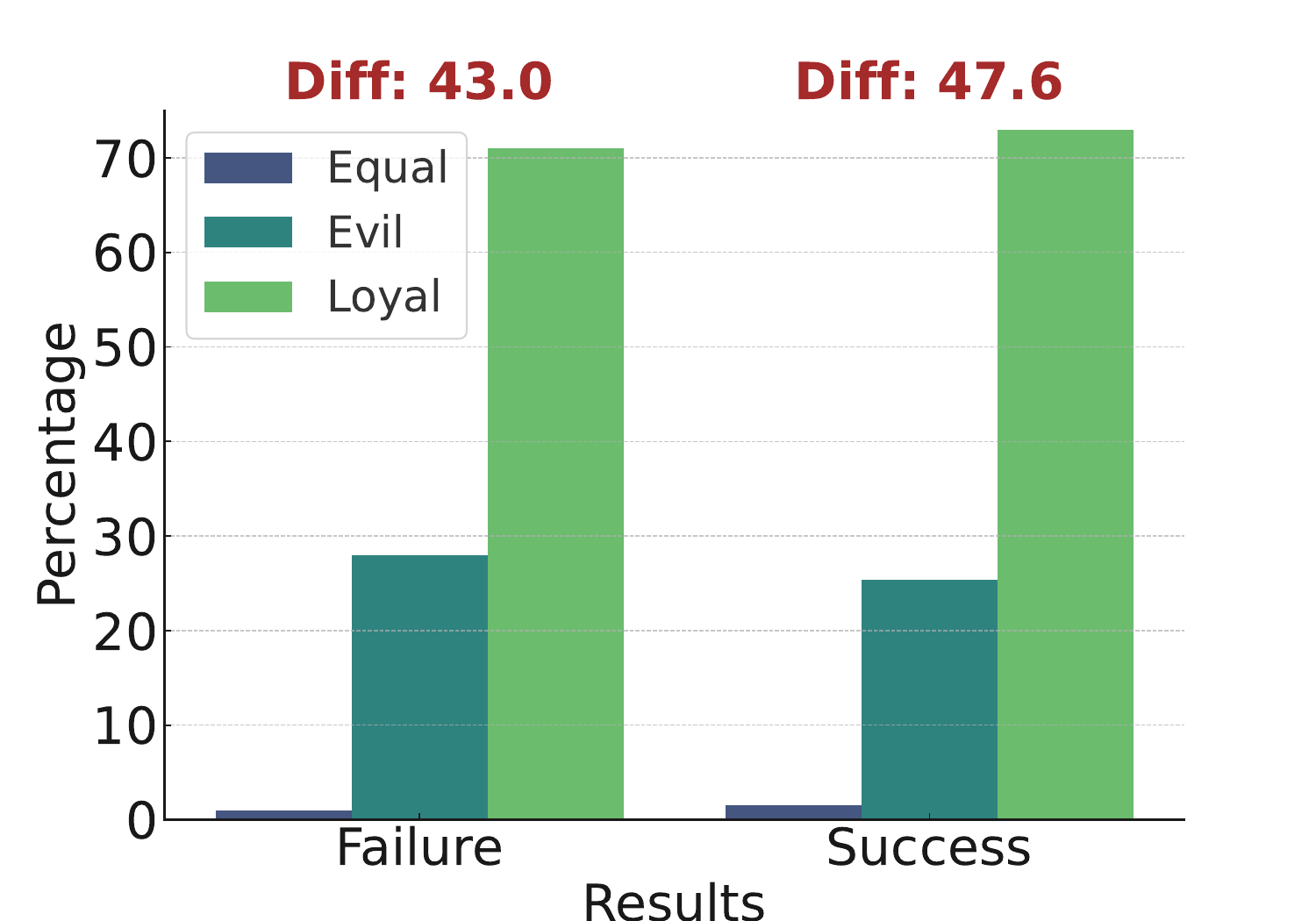}
    \caption{Intention Following $=$3 (Round Wise)}
    \label{fig:image6}
  \end{subfigure}
  \caption{Correlation between Intention Selection/Following and game performance. We present the percentages of games where evil players are equally, better, or worse than loyal players. For example, in games won by loyal players in (a), their performance matches or exceeds that of evil players. We mark the performance differences between evil and loyal players in red, showing a greater gap in successful games/quests compared to failed ones.}
  \label{fig:association_supple}
\end{figure*}

\twocolumn

\begin{table*}
\resizebox{1.0\linewidth}{!}{
\begin{tabular}{cl}
\toprule
\bf Category & \multicolumn{1}{c}{\bf Intentions}\\
\midrule
\textbf{Interrogation}
& \makecell[l]{ \textcolor{red}{Question a player on why he didn’t vote for the last team} }\\
& \makecell[l]{ \textcolor{red}{Question a player about why they changed their statements} }\\
& \makecell[l]{ \textcolor{red}{Question the leader why they selected a particular player for the team} }\\
& \makecell[l]{ Question the leader why they didn’t put themselves on the quest team }\\

\textbf{Defense}
& \makecell[l]{ Defend the proposed team composition if it includes loyal players (for Merlin)}\\
& \makecell[l]{ Defend yourself against allegations that you could be evil }\\
& \makecell[l]{ Defend your teammate against allegations that they could be evil (for evil players) }\\

\textbf{Confrontation}
& \makecell[l]{ \textcolor{red}{Share concerns about an evil player (for Merlin)} }\\
& \makecell[l]{ \textcolor{red}{Express concerns about a player from a failed quest team and suggest to remove them from current team} }\\
& \makecell[l]{ \textcolor{red}{Cast suspicion on innocent players (for evil players)} }\\
& \makecell[l]{ Counter the team proposal citing that you are good and not part of the team }\\
& \makecell[l]{ \textcolor{red}{Counter the team proposal citing that you think a player on the team is evil} }\\
& \makecell[l]{ \textcolor{red}{Express disagreement and vote disagree only if you are not in the proposed team} }\\

\textbf{Concealment}
& \makecell[l]{ Stay hidden in discussions and act like a Loyal Servant to protect yourself \\(for evil players, Merlin and Percival)}\\
& \makecell[l]{ Express that you don’t have any information on whom to put on the team to protect yourself (for Merlin) }\\
& \makecell[l]{ Pretend that you don’t have enough information about who is evil right now (for Merlin) }\\

\textbf{Deception}
& \makecell[l]{ Pretend to be Percival by telling others you are Percival (for evil/Servant)}\\
& \makecell[l]{ \textcolor{red}{Pretend to be Merlin by providing hints on who is evil (for Percival and Servant)} }\\
& \makecell[l]{ Pretend to have information and act like Merlin (for Morgana) }\\
& \makecell[l]{ \textcolor{red}{Express you are a loyal player (for evil players)} }\\
& \makecell[l]{ \textcolor{red}{Express confusion for failure of mission (for evil players)} }\\

\textbf{Persuasion}
& \makecell[l]{ \textcolor{red}{Strongly convince the team leader to choose you on team}}\\
& \makecell[l]{ Reveal identity by telling others you are Percival to persuade other loyal players(for Percival) }\\
& \makecell[l]{ \textcolor{red}{Reveal identity by telling others you are Servant to persuade other loyal players(for Servant)} }\\

\textbf{Teamwork}
& \makecell[l]{ Support team proposal}\\
& \makecell[l]{ Support one loyal player (for Merlin) }\\
& \makecell[l]{ Support your teammate to be on the quest team (for evil players) }\\
& \makecell[l]{ \textcolor{red}{Support your teammate by expressing that he is good (for evil players) }}\\
& \makecell[l]{ Include loyal players on the quest team as leader (Merlin) }\\
& \makecell[l]{ Provide hints to loyal players about evil players (Merlin) }\\
& \makecell[l]{ \textcolor{red}{Support loyal players of the previous quest team if the mission failed (for Merlin) }}\\

\bottomrule
\end{tabular}
}
\caption{Our proposed intention set. Impactful intentions are marked in red.}
\label{tab:intentions}
\end{table*}

\begin{table*}[h]
\resizebox{1.0\linewidth}{!}{
\begin{tabular}{l}
\toprule
\textbf{Structured Context - Intention Summarization}\\
\midrule
\makecell[l]{
Name: Player1\\
\\
Role: Merlin\\
\\
Role Details: \\
You are Merlin. These are your goals and characteristics: \\
- Know the identities of evil players.\\
- Subtly guide your team, especially Percival. \\
- Avoid behaviors that expose your role: overly accusing, being too helpful. \\
- Goal(Main Intent): Win without revealing identity.\\
- Player3 and Player4 are evil players\\
\\
Round: 2\\
\\
Current Leader: Player4\\
\\
Current Team: Player1, Player2, Player4\\
\\
Previous Rounds Team Voting:\\
Round 1: Player1 = disagree, Player2 = agree, Player3 = disagree, Player4 = disagree, Player5 = disagree 
\\(players didn"t agree on team)\\
...\\
\\
Previous Results: \\
Round 1: Team = Player2, Player4, Player3. Result = failure\\
\\
Previous Rounds Summary:\\
Round 1:\\
$[$summary$]$\\
...\\
\\
Previous Discussions (in the current round):\\
$[$player name$]$: $[$player's speech$]$\\
...\\
\\
Your thinking: $[$player's thinking in the current round$]$\\
\\
Your speech: $[$player's speech in the current round$]$\\
\\
Summarize your intent from your dialogues in this round.\\ 
Select multiple intents from the given options that best match your intentions from the current round.\\
Also, provide an explanation of the intents that you showed in the current round.\\
Remember this is private information to you and won't be shown to other players.\\
Remember that you can select 2-3 intents and don't use more than 50 words for explanation.\\
Intent options: $[$intent options$]$
} \\
\bottomrule
\end{tabular}
}

\caption{Example of intent summarization prompt with structured context}
\label{tab:structured_context_intent_sum}
\end{table*}

\begin{table*}
\resizebox{1.0\linewidth}{!}{
\begin{tabular}{l}
\toprule
\textbf{Structured Context - Intention Guessing}\\
\midrule
\makecell[l]{
Name: Player2\\
\\
Role: Servant\\
\\
Role Details: \\
You are a Loyal Servant of Arthur. These are your goals and characteristics: \\
- No special knowledge, rely on discussion and voting.\\
- Contribute to the success of Quests.\\
- Goal(Main Intent): Win by helping complete Quests and protecting Merlin.\\
\\
Round: 2\\
\\
Speaker Name: Player1\\
\\
Current Leader: Player4\\
\\
Current Team: Player1, Player2, Player4\\
\\
Previous Rounds Team Voting:\\
Round 1: Player1 = disagree, Player2 = agree, Player3 = disagree, Player4 = disagree, Player5 = disagree 
\\(players didn"t agree on team)\\
...\\
\\
Previous Results: \\
Round 1: Team = Player2, Player4, Player3. Result = failure\\
\\
Previous Rounds Summary:\\
Round 1:\\
$[$summary$]$\\
...\\
\\
Previous Discussions(in the current round):\\
$[$player name$]$: $[$player's speech$]$\\
...\\
\\
$[$First order prompt - Figure \ref{fig:first_prompt}$]$\\
\\
Now, here is Player1's speech: $[$player's speech$]$\\
\\
Select 2-3 intents without modifications that you think Player1 has from the given options based on your guess\\ of their role and speech.
\\
Let's think step by step before making your decisions.} \\
\bottomrule
\end{tabular}
}
\caption{Example of intent guessing prompt with structured context}
\label{tab:structured_context_intent_guess}
\end{table*}

\clearpage


\end{document}